\documentclass[preprint,12pt]{colt2024} % Include author names

\usepackage{etoc}

\usepackage{caption}
\usepackage{subcaption}

\newtheorem{assumption}[theorem]{Assumption}

\newcommand{\norm}[1]{\left\lVert#1\right\rVert}
\newcommand{\maxnorm}[1]{\left\lVert#1\right\rVert_{\mathrm{max}}}
\newcommand{\abs}[1]{\left\lvert#1\right\rvert}

\newcommand{\bbI}{\mathbb{I}}

\newcommand{\bbR}{\mathbb{R}}
\newcommand{\bbS}{\mathbb{S}}

\newcommand{\calD}{\mathcal{D}}

\newcommand{\calL}{\mathcal{L}}

\newcommand{\ba}{\boldsymbol{a}}
\newcommand{\bb}{\boldsymbol{b}}

\newcommand{\bs}{\boldsymbol{s}}

\newcommand{\bu}{\boldsymbol{u}}
\newcommand{\bv}{\boldsymbol{v}}
\newcommand{\bw}{\boldsymbol{w}}
\newcommand{\bx}{{\boldsymbol{x}}}
\newcommand{\by}{\boldsymbol{y}}

\newcommand{\bA}{\boldsymbol{A}}

\newcommand{\bC}{\boldsymbol{C}}

\newcommand{\bH}{\boldsymbol{H}}

\newcommand{\bK}{\boldsymbol{K}}

\newcommand{\bO}{\boldsymbol{O}}

\newcommand{\bV}{\boldsymbol{V}}
\newcommand{\bW}{\boldsymbol{W}}

\newcommand{\bTheta}{\mathbf{\Theta}}
\newcommand{\btheta}{\mathbf{\theta}}

\newcommand{\xpair}{\bx_{n,q}}

\newcommand{\st}{\textup{s.t.}}

\newif\ifmynotes
\mynotestrue
\usepackage{graphicx,wrapfig,lipsum}

\title[When can we approximate Wide Contrastive Models with NTKs and PCA?]{When can we Approximate Wide Contrastive Models with Neural Tangent Kernels and Principal Component Analysis?}
\usepackage{times}
\coltauthor{%
 \Name{Gautham Govind Anil} \Email{gauthamga.gga@gmail.com}\\
 \addr Indian Institute of Technology Madras
 \AND
 \Name{Pascal Esser} \Email{esser@cit.tum.de}\\
 \Name{Debarghya Ghoshdastidar} \Email{ghoshdas@cit.tum.de}\\
 \addr Technical University of Munich%
}

\begin{document}

\maketitle

\begin{abstract}%
Contrastive learning is a paradigm for learning representations from unlabelled data that has been highly successful for image and text data. Several recent works have examined contrastive losses to claim that contrastive models effectively learn spectral embeddings, while few works show relations between (wide) contrastive models and kernel principal component analysis (PCA).  However, it is not known if trained contrastive models indeed correspond to kernel methods or PCA. 
In this work, we analyze the training dynamics of two-layer contrastive models, with non-linear activation, and answer when these models are close to PCA or kernel methods. It is well known in the supervised setting that neural networks are equivalent to neural tangent kernel (NTK) machines, and that the NTK of infinitely wide networks remains constant during training. We provide the first convergence results of NTK for contrastive losses, and present a nuanced picture: NTK of wide networks remains almost constant for cosine similarity based contrastive losses, but not for losses based on dot product similarity. We further study the training dynamics of contrastive models with orthogonality constraints on output layer, which is implicitly assumed in works relating contrastive learning to spectral embedding. Our deviation bounds suggest that representations learned by contrastive models are close to the principal components of a certain matrix computed from random features. We empirically show that our theoretical results possibly hold beyond two-layer networks.

\end{abstract}


\section{Introduction}
The paradigm of self-supervised learning (SSL) builds on the idea of using knowledge about semantic similarities in the data to define which data-points should be mapped close to each other in the latent representation.
The goal of SSL is to learn a ``good representation''. While there is no unique notion of ``good'' without taking a downstream task into consideration \citep{bengio2013representation}, in general one is interested in mapping semantically similar objects to close representations in the latent space, but avoid ``dimension collapse'' that occurs when different dimensions in the latent space \emph{collapse} to the same value.
Depending on the mechanism used to prevent collapse of learned embeddings, SSL strategies can be broadly categorised as contrastive or non-contrastive learning. Contrastive learning relies on negative samples to ensure representations do not collapse \citep{oord2018representation, chen2020simple, He_2020_CVPR, haochen2021provable}, whereas non-contrastive learning avoids collapse by incorporating architectural asymmetry \citep{grill2020bootstrap, chen2021exploring} or reduction in dimension redundancy \citep{zbontar2021barlow, bardes2021vicreg}. 
In practice, a plethora of SSL strategies, including deep contrastive and non-contrastive models, have been proposed  over the past years across multiple domains; many of them demonstrating excellent performance empirically \citep{assran2022masked, wang2023masked}. %\footnote{We provide additional  related work  in Section~\ref{app: related work}.}.
While these works underline the importance of SSL and (non-)contrastive models for applications, their theoretical understanding is still limited.

Theoretical analysis of SSL is in its early stages.
There has been considerable effort in deriving generalization error bounds for downstream tasks on learned embeddings \citep{Arora2019ATA,WeiXM21,0002NN22}, and analysing spectral / isoperimetric properties of data augmentation \citep{HanYZ23,Zhuo0M023}.
Results based on learning theoretic measures \citep{saunshi2019theoretical, wei2020theoretical, nozawa2021understanding}, information theory \citep{tsai2020self, tosh2021contrastive} and loss landscapes \citep{pokle2022contrasting, ziyin2022shapes} have been studied.

Generalisation bounds, however, provide little understanding of the representations learned via SSL.
\cite{balestriero2022contrastive} answer this by showing that various (non)contrastive learning formulations result in learning spectral embedding, principal component analysis (PCA) or their variants.
In a similar vein, \cite{munkhoeva2023neural} relate contrastive learning with trace maximization problems and matrix completion---all related to PCA.
\emph{Equivalences between the optimization formulations of SSL and PCA do not necessarily imply that (non-)contrastive models, trained with gradient descent, perform PCA}.
This requires analysing either the converged solution or the training dynamics of SSL. 

A number of works derive and study the training dynamics of (non)contrastive learning,  albeit mostly limited to linear neural networks \citep{wang2020understanding, tian2021understanding, wang2021understanding, tian2022understanding, esser2023representation}.
In the context of non-linear networks, \cite{simon2023stepwise} suggest that for wide neural networks, that is, in the neural tangent kernel (NTK) regime \citep{jacot2018neural,lee2019wide}, contrastive learning could be equivalent to kernel PCA \citep{scholkopf1997kernel}.
\emph{Although no prior work explicitly analyses the convergence of wide contrastive models to kernel (or NTK) machines, there has been a significant interest in training kernel models under (non)contrastive losses} \citep{kiani2022joint, cabannes2023ssl, esser2023non}.
Depending on the problem formulation, it can indeed be shown that these kernel contrastive models are closely related to kernel PCA \citep{esser2023non} or kernel support vector machine \citep{ShahSCC22}.

\paragraph{Motivation and Contributions.}
In spite of strongly suggesting relations between constrative learning, PCA and kernel methods (or NTKs), existing theoretical works do not explicitly answer \emph{if trained contrastive models are close to kernel methods, specifically with a fixed deterministic kernel} (as has been shown in the NTK regime for supervised models). 
There is also no theoretical evidence on \emph{when trained contrastive models can be approximated by solutions of PCA or other trace maximization problems}.
We analyse the training dynamics of two-layer non-linear networks trained under contrastive or non-contrastive losses, and rigorously answer both questions. Specifically:
\begin{enumerate}
    \item In Section \ref{sec: constancy}, we derive the NTK of two-layer networks of width $M$ trained under (non)-contrastive loss, and study the deviation between NTK after several steps of gradient descent from the NTK at initialization.
    Our results address questions on the constancy of NTK.
    
    \emph{Observation 1:} (Non-)Contrastive losses are defined in terms of similarities between learned representations. We show that if the losses are in terms of dot-product similarity, then NTK drastically changes within $O(\log M)$ training time. Experiments on non-contrastive learning suggest that NTK changes \citep{simon2023stepwise}, but there was no prior theoretical evidence.
    
    \emph{Observation 2:} In contrast to dot product similarity, if the losses are defined in terms of cosine similarity---considered in \emph{InfoNCE} \citep{oord2018representation} and \emph{SimCLR} \citep{chen2020simple}---then NTK after $O(M^{1/6})$ steps is close to NTK at initialisation. Thus, contrastive models trained under such losses can be approximated by kernel methods, with a fixed NTK. Unfortunately, unlike supervised learning---where trained neural networks in NTK regime is the solution of kernel regression---there may not be a closed formed analytical solution of the trained model.

    \item In Section \ref{sec: PCA connection}, we study the training dynamics of (Grassmannian) gradient descent under orthogonality constraints of the output layer of the network.
    While orthogonality is not imposed in practical SSL approaches, it is often assumed in theoretical works to relate contrastive learning to variants of PCA \citep{munkhoeva2023neural}, in kernel SSL formulations \citep{esser2023non}, to prevent dimension collapse \citep{esser2023representation} etc. 

    \emph{Observation 1:} We note that, with orthogonality constraint, some contrastive losses (or their modifications) are equivalent to PCA of a $M\times M$ matrix $\bC(t)$ that depends on the non-linear features at the hidden layer, learned after $t$ iterations of gradient descent.

    \emph{Observation 2:} For some cosine-similarity based contrastive losses, the Frobenius norm deviation $\Vert \bC(t) - \bC(0)\Vert_F = O(t/\sqrt{M})$ suggesting that, in this case, wide contrastive models are close to PCA of a randomly initialised matrix $\bC(0)$.
    Furthermore, the representation learned via PCA from $\bC(t)$ and $\bC(0)$ are also close, upto orthonormal rotations.
\end{enumerate}
Empirical validation of our theoretical results are provided using MNIST dataset, and we further show that some of the results may also hold beyond two-layer networks (see Section~\ref{sec:discussion}). All proofs are provided in the supplementary material.

\section{Preliminaries and Problem Setup}\label{sec: setup}
Before going into the main results of the paper, we first outline the contrastive learning setup, the embedding function and NTK formulation under consideration,  together with the general conditions for the NTK to remain constant during training. We use the following notation throughout the paper:

\paragraph{Notation.} We use lowercase bold letters (e.g. $\ba$) to denote vectors and upper case bold letters (e.g. $\bA$) to denote matrices. Let $\bA_{i.}$ denote the $i^{th}$ row and $\bA_{.i}$ denote the $i^{th}$ column of matrix $\bA$. Let $\bbI$ be an appropriately sized identity matrix. $\norm{ \cdot }_p$ denotes the $L_p$ norm, $\norm{ \cdot }_F$ denotes the Frobenius norm and $\maxnorm{\bA}: = \max_{ij} \left \{\abs{\bA_{ij}} \right\}$. We denote parameter $\bTheta$ at time-step $t$ by $\bTheta(t)$; however the time indexing is suppressed when it is clear from the context to improve readability.

\subsection{(Non-)Contrastive Learning}

In this work, our primary focus is  on sample-contrastive methods which use multiple positive/ negative sample pairs. 
Consider a dataset of $N$ datapoints: $\calD:=\big\{\big\{ \bx_n,\xpair\big\}_{q=1}^Q\big\}_{n=1}^N$, where $\bx_n\in\bbR^D$ denotes the $n^{th}$ $D$ dimensional data sample and $\xpair $ denotes the $q^{th}$ pair in relation to $\bx_n$.\footnote{Note that the pair could involve a positive or negative sample. Hence, this framework encompasses popular examples such as the \emph{contrastive triplet} setting $\{\bx_n,\bx_n^+,\bx_n^-\}_{n=1}^N$ and the \emph{non-contrastive} setting $\{\bx_n,\bx_n^+\}_{n=1}^N$ .}  Using this formulation, we now state a general form for the contrastive loss:
\begin{equation}
    \label{eqn:loss}
\calL(\calD) := \frac{1}{N} \sum_{n = 1}^{N} l \left(\left\{ s\left(\bx_n,\xpair \right)\right\}_{q=1}^Q\right)
\end{equation}
where $l( \cdot )$ is some function and $s(\bx,\tilde\bx )$ is the similarity between representations of inputs $\bx$ and $\tilde\bx$ learned by a (non-)contrastive model. 
%and $l( \cdot )$ is some function of  $s(\bx,\tilde\bx )$. 
While \emph{softmax} or its logarithm are typically used for $l(\cdot)$ in practice, theoretical works often consider $l(\cdot)$ to be \emph{linear} \citep{ji2023power, esser2023representation}. %The widely used contrastive loss SimCLR reduces to the case of $l(\cdot)$ being linear. 
While a wide range of similarity measures $s(\cdot,\cdot)$ are considered, they often build on similar underlying ideas. Losses such as \emph{MoCo} \citep{He_2020_CVPR} build on \emph{dot product similarity}, while the popular \emph{SimCLR}  and \emph{InfoNCE} \citep{chen2020simple, oord2018representation} losses build on \emph{cosine similarity}. Therefore, we consider the following two similarity measures,  where $\bx \mapsto f(\bx)$ denotes the learned representation:\footnote{Note that $f(\cdot)$ is a parameterized function as we later define in \eqref{eqn: ntk_nn}. However, we suppress the parameterization here for ease of notation.}
\begin{align}
    s({\bx, \tilde\bx }) &:= f(\bx )^\top f({\tilde\bx }),\tag{dot product}\label{eq:dot}\\
    s({\bx, \tilde\bx }) &:= \frac{f(\bx )^\top f({\tilde\bx })}{(\norm{f(\bx )} + \delta)(\norm{f({\tilde\bx })} + \delta)}. \tag{cosine similarity}\label{eq:cos}
\end{align}
We consider the following set of assumptions on the similarity measure and on the data:
\begin{assumption}[Constant for cosine similarity]
    \label{assm:delta}
    $\delta$ is a small strictly positive constant.
\end{assumption}
\begin{assumption}[Smoothness]
\label{assm:smooth}
$\abs{\frac{\partial l(~\cdot~)}{\partial s(\bx , {\tilde\bx })}} \leq c_l$ for any $\bx , {\tilde\bx }$. 
\end{assumption}
\begin{assumption}[Bounded inputs]
    \label{assm:data}
    Input vectors are bounded, $\max_{n,q} \left\{\norm{{\bx_n}}_{\infty},\norm{{\bx_{n,q}}}_{\infty} \right\} \leq c_{in}$.
\end{assumption}
While $\delta$ is not typically considered in cosine similarity, assuming $\delta>0$ ensures that $s({\bx, \tilde\bx })$ is defined even when norms of the representations are zero. Furthermore, $\delta>0$ can be made arbitrarily small, making Assumption~\ref{assm:delta} practically reasonable. Apart from making the cosine similarity computation numerically stable, this structure for cosine similarity helps to simplify the proofs by providing a strictly positive lower bound on $\norm{f(\bx )} + \delta$ for any $\bx$. Assumption~\ref{assm:smooth} is evidently satisfied for commonly considered losses where $l(\cdot)$ is \emph{linear} or \emph{softmax} . Assumption~\ref{assm:data} is often considered for theoretical analysis in NTK literature (e.g. \cite{jacot2018neural}).
% Assumptions~\ref{assm:smooth}~\&~\ref{assm:data}: Smoothness assumptions on the loss and parts of the network as well as boundedness of inputs as well as weights are common in the context of similar types of analysis in the supervised setting (e.g. \cite{jacot2018neural}). 

\subsection{Embedding Function} 
The outlined setup for contrastive losses is stated for an arbitrary embedding function $f(\cdot)$. However, for our analysis, we focus on one hidden layer neural networks. We aim to find a mapping $ f(\bx; \btheta):\bbR^D\rightarrow\bbR^Z$ parameterized by $\btheta$ where typically $D>Z$. In particular, we consider a two-layer fully connected non-linear neural network:
$ f(\bx; \btheta) = \bW^\top\phi(\bV\bx) $
where $\bx\in\bbR^D$ is an input vector and $\phi$ is a pointwise non-linear activation function. $\bW\in\bbR^{M\times D}$ and $\bV\in\bbR^{M\times Z}$ are the trainable weight matrices and let $\btheta$ be the vector which contains all entries of $\bW$ and $\bV$. 
In the context of (infinite) width analysis, the `appropriate' initialization of these weights is essential.
Existing NTK literature on supervised learning (e.g. \cite{jacot2018neural,arora2019exact}) considers the following parameterization:
\begin{equation}
    \label{eqn: ntk_nn}
    f(\bx; \btheta) := \frac{1}{\sqrt{M}}\bW^\top\phi(\bV\bx)%, \quad \bW_{i,j},\bV_{i,j}\sim \mathcal{N}(0, {1})\forall i\in[M],j\in[D]
\end{equation}
where each $\bW_{i,j},\bV_{i,j}\sim \mathcal{N}(0, {1})$.
We consider this setup, termed  \emph{NTK parametrization}, for the remainder of the paper.
In addition, we also consider the following assumptions:
\begin{assumption}[Max norm of weights at initialization]
    \label{assm:w_init}
    $\maxnorm{\bW(0)}$,  $\maxnorm{\bV(0)} \leq c_\btheta \log M$.
    % $\norm{\textbf{W$_1$}(0)}_{\infty}, \norm{\textbf{W$_2$}(0)}_{\infty} \leq c_w$.
\end{assumption}
%\begin{assumption}[Spectral norm of weights at initialization]
%    \label{assm:w_spec}
%    $\norm{\bW(0)}_2$,  $\norm{\bV(0)}_2 \leq c_s \sqrt{M}$.
    % $\norm{\textbf{W$_1$}(0)}_{\infty}, \norm{\textbf{W$_2$}(0)}_{\infty} \leq c_w$.
%\end{assumption}

\begin{assumption}[Smoothness of activation function]
    \label{assm:activation}
    $\phi$ is $L_{\phi}$-Lipschitz and $\beta_{\phi}$-smooth.  
\end{assumption}
\begin{assumption}[Bounds on gradients and weights]
    \label{assm:gradient} Let $\bb_1 = \frac{\partial f}{\partial \bx}$,  $\bb_2 = \frac{\partial f}{\partial \phi( \bV\bx)}$. At initialization, $\norm{\bW(0)}_2$,  $\norm{\bV(0)}_2 \leq c_s \sqrt{M}$ and there is constant $s_0$ such that $\norm{\bb_i}_{\infty} \leq \frac{s_0}{\sqrt{M}}\norm{\bb_i}_2$ \; for \; $i = 1, 2$. 
  % \note{TODO add exact form}
\end{assumption}
Assumption~\ref{assm:w_init} holds with high probability for standard Gaussian initialisation of weights. Assumption~\ref{assm:activation} is a usually considered smoothness criterion (and holds for sigmoid, tanh etc.).
Assumption~\ref{assm:gradient} is typically needed to prove constancy of NTK for supervised models \citep[see][]{liu2020toward,liu2020linearity}.

\subsection{Conditions for Constancy of NTK} 

Let us start by outlining the NTK analysis in general for a function $f(\bx;\btheta(t)):\bbR^D\rightarrow\bbR^Z$, where $Z\geq 1$.
For input vectors $\bx\in\bbR^D$ and $\tilde\bx \in\bbR^D$, we define the \emph{empirical NTK} for a neural network $f(\cdot)$ parameterized by $\btheta(t)$ as:
\begin{align*}
    \bK_{ij}(\bx , {\tilde\bx }; \btheta(t)) := {\frac{\partial f_i(\bx ;\btheta(t))}{\partial \btheta(t)}}^\top\frac{\partial f_j(\tilde\bx ;\btheta(t))}{\partial \btheta(t)}
\end{align*}
where $f_i(\bx ;\btheta)$ represent the $i^{\text{th}}$  entry of the $Z$ dimensional function output.
In general, $ \bK(\bx , {\tilde\bx }; \btheta(t))$ varies with time $t$ as the model is trained. However, under certain conditions, for infinitely wide neural networks, the NTK stays constant during training \citep{jacot2018neural,arora2019exact}, i.e,
  \begin{align}\label{eq:NTK constant}
         \forall t \quad \abs{\bK_{ij}(\bx , {\tilde\bx }; \btheta(t)) - \bK_{ij}(\bx , {\tilde\bx }; \btheta(0))} \rightarrow 0 \text{ as } M\to\infty.
        \end{align}
Furthermore, under Gaussian initialisation of parameters, it holds that the NTK at initialisation converges, as $M\to\infty$, to an  \emph{analytical NTK} 
%$\bK^*(\bx,\tilde\bx )$, which is computed as the expectation over the weights at initialization:
% \begin{align*}
  $ \bK_{ij}^*(\bx,\tilde\bx ) := \mathbb{E}_{\btheta} \left[   \bK_{ij}(\bx , {\tilde\bx }; \btheta)\right]$.
% \end{align*}
As the NTK does not change during training, the training dynamics of the network at any time step can be written in terms of $\bK^*$ --- in the supervised setting, this leads to kernel regression at convergence.
To prove constancy of the form \eqref{eq:NTK constant}, several works have analyzed NTKs in the supervised setting \citep{arora2019exact, lee2019wide, chizat2019lazy}. In particular, \cite{liu2020linearity} showed that the constancy of NTK is predicated on the spectral norm of the Hessian.

To study the NTK of contrastive models,  we consider $f(\bx;\btheta)$ of the form \eqref{eqn: ntk_nn} and use the machinery built in \citep{liu2020linearity, liu2020toward}.
Define $Z$ Hessian matrices, one for each element of the output representation. The $z^{\text{th}}$ Hessian matrix $\bH^{(z)}$ (evaluated at input $\bx$) is
% \begin{equation*}
 $   \bH_{ij}^{(z)}(\bx) :=  \frac{\partial^2 f_z(\bx;\btheta(t))}{\partial \btheta_{i}(t) \partial \btheta_{j}(t)},\ 
     z\in[Z].$
% \end{equation*}
We bound the change in the spectral norm of the Hessian in terms of the change in weights by adapting \citet[][Theorem 7.1]{liu2020toward}\footnote{Theorem 7.1 of \citep{liu2020toward} gives a bound of the form $\norm{\bH^{(z)}(\bx ; \btheta(t))}_2 = O\left(\frac{R^{3L}}{\sqrt{M}}\right)$ for a network with $L$ layers. However, for two-layer networks, it is possible to reduce this bound to the form given in Lemma \ref{lemma:hessian}.} to account for multi-dimensional outputs:
\begin{lemma}[Bound on the norm of the Hessian]
    \label{lemma:hessian}
     Under Assumptions~\ref{assm:data}, \ref{assm:activation}, \ref{assm:gradient}, consider the neural network defined in \eqref{eqn: ntk_nn}. If the change in weights during training is bounded as 
     \begin{align}
         \norm{\bW(t) - \bW(0)}_F + \norm{\bV(t) - \bV(0)}_F \leq R,
         \label{eq: R bound}
         \end{align}
          then, $\forall \; z \in [Z]$, with $\alpha_1 = 4\beta_{\phi} c_{in}^2 L_{\phi}$ and $\alpha_2 = 4L_{\phi}c_{in}(1 + \beta_{\phi} c_{in}s_0 c_s)$,  the $z^\text{th}$ Hessian is bounded as:
   $  \norm{\bH^{(z)}(\bx ; \btheta(t))}_2 \leq \frac{\alpha_1 R + \alpha_2}{\sqrt{M}}.$
\end{lemma}
 With help of Lemma \ref{lemma:hessian}, we can now bound the change in NTK. Towards this, we extend Proposition~2.3 of \cite{liu2020linearity} to the multi-dimensional case $(Z>1)$ to obtain the following lemma:
\begin{lemma}[Bound on the change in NTK]
    \label{lemma:ntkhess}
   Define $\bbS := \{\bs \in \mathbb{R}^p; \norm{\bs - \bs(0)} \leq R\}$, where $p$ is the total number of learnable parameters in \eqref{eqn: ntk_nn}. Assume that for any input $\bx$, $\norm{\bH^{(z)}(\bx; \bs)}_2 \leq \epsilon$ and $\norm{\nabla_{\bs}f_z({\bx; {\bs}})}_2 \leq c_0$, $\forall \; z \in [Z]$ and $\forall \; \bs \in \bbS$. Then, for any inputs $\bx, \tilde\bx$, $\forall \; \bs \in \bbS$ and $\forall \; i,j \in [Z]$, $\abs{\bK_{ij}(\bx , {\tilde\bx }; {\bs}) - \bK_{ij}(\bx , {\tilde\bx }; {\bs(0)})} \leq 2\epsilon c_0 R$.
\end{lemma}
If $\bK$ does not change during training, the analytical (expected) NTK $\bK^*$ models the behaviour of the network not only  at initialization, but also at convergence and therefore allows us to express the network dynamics in a simple form.
In supervised settings with squared loss, it is known that condition \eqref{eq: R bound} is true till convergence for wide neural networks \citep{liu2020toward}.
While it is possible to use Lemmas \ref{lemma:hessian} and \ref{lemma:ntkhess} to examine the behaviour of NTK in general, note that it is \textbf{not known when the condition \eqref{eq: R bound} holds for an arbitrary loss function}.  In Section \ref{sec: constancy}, we study the validity of this condition when learning embeddings using aforementioned (non-)contrastive losses.  

\section{On the Constancy of NTKs under Contrastive Losses}\label{sec: constancy}

We now examine the NTK evolution for neural networks trained under contrastive losses. To do so,
using the above presented setup, we derive the dynamics of a neural network trained using gradient flow under a loss of the form (\ref{eqn:loss}) in terms of the NTK of the neural network as defined in \eqref{eqn: ntk_nn}:
\begin{lemma}[Contrastive learning dynamics in terms of NTK]
\label{lemma:ntk_dyn}
    % Under Assumptions~\ref{assm:}~-~\ref{assm:} \note{add}
    % \note{ gradient descent or gradient flow?} 
    Consider training a neural network of the form (\ref{eqn: ntk_nn}) using a loss $l(\cdot)$ of the form (\ref{eqn:loss}) under gradient flow on dataset $\mathcal{D}$. Let  $g_i(\bx, \tilde\bx;\btheta(t)  ) := \frac{\partial s(\bx, \tilde\bx  )}{\partial f_i(\bx;\btheta(t))}$. Then, for $z \in [Z]$, the representation of an arbitrary input $\tilde\bx $ evolves as:
    \begin{align*}
    \frac{\partial {f}_z({\tilde\bx;\btheta(t) })}{\partial t}=
    -\frac{1}{N}\sum_{n, q} \frac{\partial l(\cdot)}{\partial s(\bx_n, \xpair )}
         &\left[ \sum_{i=1}^{Z} \left[\bK_{zi}(\tilde\bx  , \bx_n; \btheta(t))g_i(\bx_n,\xpair;\btheta(t) )  \right.\right.\\
         &\left.\qquad\qquad+ \bK_{zi}(\tilde\bx  , \xpair ; \btheta(t))g_i(\xpair , \bx_n;\btheta(t)) \right]\big].
    \end{align*}
    % where $g_i(\textbf{x}, \tilde\bx  ) := \frac{\partial s(\textbf{x}, \tilde\bx  )}{\partial f_i(\bx)}$ and $z \in [Z]$.
\end{lemma}
While Lemma~\ref{lemma:ntk_dyn} holds for any loss of the form \eqref{eqn:loss}, we are interested in the behaviour of the NTK when trained under losses which use \ref{eq:dot} or \ref{eq:cos} as the similarity measure.
\begin{wrapfigure}[11]{r}{5.5cm}
\vspace{-10pt}
\includegraphics[width=5.5cm]{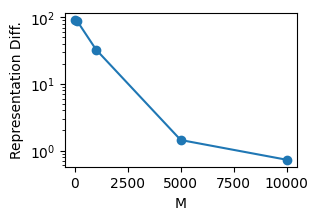}
\vspace{-25pt}
\caption{Difference between NTK and GD dynamics.}\label{fig:dynamics}
\end{wrapfigure} 

\paragraph{Numerical Simulation.}
Throughout the paper, we illustrate our theoretical findings numerically. If not otherwise stated, all experiments are performed on $1000$ randomly sampled points of the MNIST dataset \citep{deng2012mnist}\footnote{Note that increasing number of samples would not have an effect on the overall trends due to the presence of the $\frac{1}{N}$ term in the loss as defined in (\ref{eqn:loss}).}. The positive samples are obtained by randomly resizing and cropping the corresponding data sample. Negative samples are obtained by randomly sampling from the entire dataset. For all experiments, we generate one positive sample and one negative sample for each data sample. All plots indicate the average over $5$ runs and additional plots are provided in the supplementary material. If not otherwise stated, we consider ReLU activation function and linear contrastive loss. Before going into theoretical analysis, we empirically illustrate that the dynamics derived from the NTK (at initialization) is similar to the actual dynamics obtained by training the network. We first train \eqref{eqn: ntk_nn} using cosine similarity for $500$ epochs. We repeat the same using the NTK dynamics in Lemma~\ref{lemma:ntk_dyn}. For this simulation, we consider $N=Z=10$. In Figure~\ref{fig:dynamics}, we plot the fractional difference between the representations learned by the two methods after taking the maximum across inputs and dimensions. We observe that as $M$ increases, the difference decreases and the dynamics of Lemma~\ref{lemma:ntk_dyn} align with training \eqref{eqn: ntk_nn} using gradient descent.

\subsection{NTK for Dot product Losses Does Not Necessarily Stay Constant}\label{sec: dot product}
We start with examining the change in weights (as defined in  \eqref{eq: R bound}) of a network trained using contrastive losses which utilize \ref{eq:dot} as the similarity measure. In particular, we show that in this setting, there exists cases where the change in weights become arbitrarily large even for arbitrarily wide neural networks, implying that the NTK does not remain constant. To demonstrate this, we consider a simple loss function of the form \eqref{eqn:loss} under \ref{eq:dot} similarity.
Because there is no normalization in the dot product similarity measure, the loss can be minimized arbitrarily by scaling the weights and hence we expect the weights to grow arbitrarily large with time. We formalize this notion and its implications on the constancy of the NTK  in the following proposition:
\begin{proposition}[NTK under dot product does not remain constant]
    \label{prop:lin_change}
    % Under  Assumptions~\ref{assm:}~-~\ref{assm:} \note{add}  and
    For $D=Z=1$, linear loss $(l(a):=a)$, \ref{eq:dot} similarity and triplet setting $(\calD = \{x_n,x_n^+,x_n^-\}_{n=1}^N)$ in \eqref{eqn:loss}, the optimization is:
    % \begin{align*}
    $
         \min_\btheta   \frac{1}{N}\sum_{n = 1}^N  f(x_n;\btheta)\left(f(x_n^-;\btheta) - f(x_n^+;\btheta)\right)$.
    % \end{align*}
    % . Assume a linear network $(\phi(a):=a)$ and linear loss $(l(a):=a)$, trained under gradient flow. for a triplet setting $\calD = \{x_n,x_n^+,x_n^-\}_{n=1}^N$ can be written as 
    % Assume that ${\bw}^\top {\bv}$ is non-zero at initialization. Consider $ f_{lin}(\cdot)$, trained using the loss \eqref{eqn:lin_loss} (which is a \ref{eq:dot} loss) using gradient flow. 
    Consider a network \eqref{eqn: ntk_nn} with linear activation $(\phi(a):=a)$, weights initialised as independent $\mathcal{N}(0,1)$, and trained via gradient flow.
    \\
    There is a dataset such that, with probability at least  $1 - \frac{25}{\sqrt{M}}$, for a time step $\Tilde{t} \in (0, \log M)$ and any input pair $x,\Tilde{x}$ with $x\Tilde{x} \neq 0$, the NTK satisfies 
      $    \abs{\bK(x, \tilde x; \btheta(t)) -  \bK(x, \tilde x; \btheta(0))} \rightarrow \infty$ as $t \rightarrow \Tilde{t}$.
        % \end{align*}
\end{proposition}
Proposition \ref{prop:lin_change} shows there are cases where the \emph{NTK does not remain constant} even for arbitrarily wide networks and logarithmic training time when trained under dot product similarity based loss.

\begin{wrapfigure}[8]{r}{5.5cm}
\vspace{-10pt}
\includegraphics[width=5.5cm]{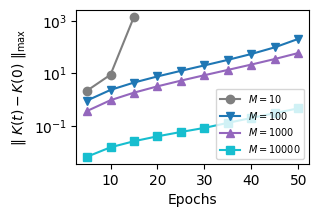}
\vspace{-22pt}
\caption{Change in NTK.}\label{fig:dotproduct}
\end{wrapfigure} 
\paragraph{Numerical Simulation on MNIST.} To show that Proposition~\ref{prop:lin_change} is true in practice, we look at the deviation of the NTK with training across networks of varying widths. The results are shown in Figure~\ref{fig:dotproduct}. We observe that, as expected, even after a few epochs, the difference in NTK diverges.

\subsection{NTK for Cosine similarity Losses Remains Constant}\label{sec: cosine smilarity}
Considering the same question as in the previous section, we now shift our focus onto the constancy of NTK for losses defined in terms of the \ref{eq:cos}. The key difference between dot product and cosine similarity is the presence of \emph{normalization by norms of the representations}. We now show that this normalization plays an important role in deciding the learning dynamics and examine its implications on the constancy of the NTK.
To prove constancy of the NTK, we make use of the fact that the similarity measure is normalized and first establish a bound on the maximum element-wise change in weights.
\begin{lemma}[Bound on element-wise change in weights under cosine similarity]
    \label{lemma:change}
    Under  Assumptions~\ref{assm:delta}~-~\ref{assm:activation}, consider losses of the form \eqref{eqn:loss} where \ref{eq:cos} is used. If a neural network $f(\cdot)$ as defined in \eqref{eqn: ntk_nn} is trained using gradient descent with learning rate $\eta$, at any time $t$, the change in weights are bounded as:
 $   \abs{\Delta \bV_{ij}(t)} \leq \frac{\beta_1}{\sqrt{M}} \; %\max_{ij} \left \{\abs{\bV_{ij}(t)} \right\}
    \maxnorm{\bW(t)}
    $ and $
 \abs{\Delta \bW_{ij}(t)} \leq \frac{\beta_2}{\sqrt{M}} \; %\max_{ij} \left \{\abs{\bW_{ij}(t)} \right\} 
 \maxnorm{\bV(t)}$
    % \end{align*}
    where 
$\beta_1 =\frac{4}{ \delta} c_lc_{in}Q\sqrt{Z} L_{\phi} $ and $\beta_2 = \frac{4}{ \delta} c_lc_{in}QDL_{\phi}$ are constants independent of $M$.
\end{lemma}
From Lemma~\ref{lemma:change}, it can be seen that bounds for change in $\bV$ and $\bW$ form a coupled system. To study the discrete-time dynamics of this system, we define and characterize a useful quantity $c(t)$:
\begin{lemma}[Bound on weight difference during training under cosine similarity]
    \label{lemma:ct}
    Define $\beta := \max \{\beta_1, \beta_2\}$ and 
    $ c(t) := c(0)\left(1 + \frac{\beta}{\sqrt{M}} \right)^t $
    where $c(0) = c_{\theta} \log M$.
    % $ c(0) := \max\left\{\maxnorm{\bV(0)},\maxnorm{\bW(0)}\right\}$.
    Then, for any $t$, we have :
    % \begin{align*}
  $   \maxnorm{\bV(t) - \bV(0)} \leq  c(t) - c(0) $ and $
    \maxnorm{\bW(t) - \bW(0)} \leq  c(t) - c(0). $
     % \end{align*}
\end{lemma}
We now state the main theorem regarding the convergence of NTK for cosine similarity losses:
\begin{theorem}[Bound on the change in NTK under cosine similarity]
    \label{thm:ntk_conv}
     Consider losses of the form \eqref{eqn:loss} with \ref{eq:cos}. 
     Let $ c(0),\beta$ be the constant in Lemma~\ref{lemma:ct},   $R$ be as in \eqref{eq: R bound}\footnote{{Note that $R$ here is a function of $M$, with the relation being given by Lemma \ref{lemma:ct}. Similarly, $c(0) = c_{\theta} \log M$.}}, $\alpha_1, \alpha_2$ be as in Lemma \ref{lemma:hessian} and $\gamma:=2\sqrt{2}DL_{\phi}c_{in}$. If a neural network $f(\cdot)$ of the form \eqref{eqn: ntk_nn} is trained using gradient descent, then under Assumptions~\ref{assm:delta}~-~\ref{assm:gradient}, for $t \leq M^{\alpha}$ iterations, the change in NTK is bounded as 
     \begin{align*}
         \abs{\bK_{ij}(\bx , {\tilde\bx }; \btheta(t)) - \bK_{ij}(\bx , {\tilde\bx }; \btheta(0))} \leq \gamma  \left(c(0) e^{{\beta}{{M^{\alpha - 0.5}}}} \right) \frac{\alpha_1R^2 + \alpha_2R}{\sqrt{M}} .
        \end{align*}
    In particular, if we set $\alpha = \frac{1}{6}$ and assume $M \geq \max\{1, \beta^3\}$, then the above statement simplifies to
    \begin{align*}
        \max_{t \in \left(0, M^{1/6}\right]}\  \sup_{\bx, \Tilde{\bx}}\  \abs{\bK_{ij}(\bx , {\tilde\bx }; \btheta(t)) - \bK_{ij}(\bx , {\tilde\bx }; \btheta(0))} = O\left(M^{-1/6}(\log M)^3 \right).
    \end{align*}
\end{theorem}
%We can now use Theorem~\ref{thm:ntk_conv} to state the limiting behaviour of the NTK as the width goes to infinity and obtain a result of the form \eqref{eq:NTK constant} for the contrastive setting:
%\begin{corollary}[Constancy of NTK under cosine similarity]
%\label{cor:ntk_conv_infty}
%In the setting of Theorem~\ref{thm:ntk_conv}, let $\alpha < \frac{1}{4}$. Then, for any fixed $t$, 
% \begin{align*}
%   $ \abs{\bK_{ij}(\bx , {\tilde\bx }; \btheta(t)) - \bK_{ij}(\bx , {\tilde\bx }; \btheta(0))}\rightarrow0$  as $M \rightarrow \infty$.
% \end{align*}
%\end{corollary}
According to Theorem \ref{thm:ntk_conv}, wide neural networks trained under \ref{eq:cos} based contrastive loss have a nearly constant NTK even after $M^{1/6}$ iterations of gradient descent. This is in sharp contrast to networks trained under \ref{eq:dot} based losses where the change in weights become arbitrarily large within $\log M$ gradient descent updates. Intuitively, this holds since normalization ensures that the change in weights remain sufficiently small in the case of \ref{eq:cos}.

\begin{wrapfigure}[8]{r}{5.5cm}
\vspace{-20pt}
\includegraphics[width=5.4cm]{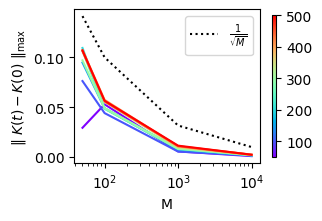}
\vspace{-15pt}
\caption{Change in NTK.}
\label{fig:1layer}
\end{wrapfigure} 
\paragraph{Numerical Simulation on MNIST.}
To check the validity of Theorem~\ref{thm:ntk_conv}, we look at the the change in NTK with training across networks of varying widths. The results are shown in Figure~\ref{fig:1layer}. In general, we expect the change in each entry of the NTK to decrease with an increase in depth. We see that this is indeed the case. For a fixed number of iterations, $t= O(1)$, we expect $\alpha\approx 0$ and the change to go down (roughly) as $1/\sqrt{M}$ with increasing width.

\begin{remark}[Closed form solution in terms of NTK]\label{rem:closed form}
% In the supervised setting, taking advantage of the constancy results, the NTK analysis directly offers a closed form representation of the neural network and therefore a way to perform inference through Kernel regression \citep{jacot2018neural}.
In the supervised setting, combining the analytical NTK with the closed form solution for kernel regression provides a closed form expression of the network trained until convergence \citep{jacot2018neural}, or even when early stopped.
We take a step towards such a result for contrastive losses. In Lemma~\ref{lemma:ntk_dyn}, we present the learning dynamics in the contrastive loss setting in terms of the NTK and in Theorem~\ref{thm:ntk_conv}, we show that the NTK remains constant during training for $M^{1/6}$ steps under \ref{eq:cos}. While this is an important step towards a better theoretical understanding of contrastive models, it does not yet provide a closed form solution of the model output in terms of the NTK.
While prior works on kernel contrastive methods suggest that, in the wide neural network regime, contrastive losses could be equivalent to kernel PCA \citep{simon2023stepwise}, this connection has not been proven so far and is not apparent from the dynamics derived in Lemma~\ref{lemma:ntk_dyn} even if the NTK is constant (for cosine similarity based losses). Therefore, we shift our viewpoint from NTK dynamics to explicitly investigating this connection.
\end{remark}

\section{On the Connection between Wide Contrastive Models and PCA}\label{sec: PCA connection}
We next study if there is a formal connection between PCA and representations learned by contrastive models, trained with gradient descent. Prior works have connected contrastive models to a trace maximization problem with an orthogonality constraint on the output layer, of the form:
  \begin{align}\label{eq: PCA}
    \max_{\bW,\vartheta} \text{Tr}\left( \bW^\top \bC_\vartheta\bW  \right) \quad\st\quad \bW^\top \bW = \bbI_Z,
\end{align}
where $\bC_\vartheta\in\bbR^{M\times M}$ is a symmetric matrix that has a (possibly non-linear) data dependence through a function parameterized by $\vartheta$. If $\bC_\vartheta$ stays constant during optimization, optimization is done only over $\bW$ and hence the problem simplifies to PCA on $\bC_\vartheta$. To connect contrastive losses and PCA, it is then necessary to analyze the behaviour of $\bC_\vartheta$ when a contrastive model is trained. Existing works do not examine this aspect of neural network dynamics.
% In this assumption lies the challenge of connecting contrastive losses and PCA solutions.
More specifically, \cite{esser2023non} considers a kernel setting where learning is done using a contrastive loss of the form \eqref{eq: PCA}, but does not link it to the neural network dynamics. \cite{esser2023representation} considers neural network dynamics for contrastive losses of the form \eqref{eq: PCA} but only examines the linear setting. \cite{simon2023stepwise} considers the dynamics for kernel models but does not investigate if the kernel dynamics are close to the neural network dynamics. \cite{munkhoeva2023neural} reformulates losses such as \emph{SimCLR} to \eqref{eq: PCA}.%, but does not consider the implications of the reformulation on the learning dynamics. 

Extending prior works, we work towards formalizing the connection between \textbf{non-linear, wide} networks and PCA, not only through rewriting the loss, but also by taking \textbf{learning into consideration}.
Let us consider \eqref{eqn:loss} with a linear loss such that we obtain \eqref{eqn:orth_w2}. In addition, we consider \eqref{eqn: ntk_nn} under orthogonality constraint on the second layer to obtain a neural network of the form \eqref{eq:orth_nn}:
  \begin{align}
&\calL(\calD) := - \frac{1}{N}\sum^N_{n=1} \sum^Q_{q = 1} {\alpha_q} s(\bx_n,\bx_{nq}) \label{eqn:orth_w2}\\
&f^{\perp}(\bx; \btheta) := \frac{1}{\sqrt{M}}\bW^\top\phi(\bV\bx)  \quad \st \quad \bW^\top \bW = \bbI_Z.\label{eq:orth_nn}
\end{align}
%     \begin{align}
% &\calL^{\textrm{cos}}(\calD) := \frac{1}{N}\sum^N_{n=1} \sum^Q_{q = 1}\left( {\alpha_q} s(\bx_n,\bx_{nq})\right)\quad\text{with \ref{eq:cos} and} \label{eqn:orth_w2}\\
% &  f^{\perp}(\bx; \btheta) := \frac{1}{\sqrt{M}}\bW^\top\phi(\bV\bx)  \quad \st \quad \bW^\top \bW = \bbI_Z.\label{eq:orth_nn}
% \end{align}
% where again $\bx_i^-$ and  $\bx_i^+$ denote the positive and negative samples corresponding to the  $\bx_i$. 
where $\alpha_q = 1$ if $(\bx_n,\bx_{nq})$ is a positive pair and $\alpha_q = -1$ for a negative pair. 
Observe that if we use the \ref{eq:dot} similarity, $s(\bx,\tilde\bx) = f^{\perp}(\bx;\btheta )^\top f^{\perp}({\tilde\bx;\btheta })$, then the minimisation of the contrastive loss $\calL(\calD)$ in \eqref{eqn:orth_w2} can be directly posed as a trace maximization problem\footnote{The value of the Trace in \eqref{eq: PCA} is the same irrespective of whether we use $\bC_V$ or $\widetilde\bC_V$. However, using the symmetric $\widetilde\bC_V$ allows us to make the connection of solving \eqref{eq: PCA} through PCA more direct.} 
\begin{equation}
\label{eqn:lin_pca}
    \eqref{eq: PCA} \text{ with }
    \bC_\vartheta=\widetilde\bC_V=\frac{\bC_V + \bC_V^\top}{2};~ 
  \bC_V = \frac{1}{MN} \sum_{n = 1}^N \sum_{q = 1}^Q  
{\alpha_q\phi({\bV\bx_{n,q}})\phi({\bV\bx_n})^\top}. 
\end{equation}
\begin{remark}[From contrastive models to PCA in case of dot product similarity]
    Proposition~\ref{prop:lin_change} shows that the NTK diverges within $\log(M)$ steps for training with \ref{eq:dot} based losses.
    While we do not explicitly prove this, it also follows that, as the first layer $\bV$ is trained, the matrix $\bC_{\bV}(t)$ diverges arbitrarily from the initialisation $\bC_{\bV}(0)$.
    Hence, while minimizing $\calL(\calD)$ for any fixed $\bV$ corresponds to PCA for finding $\bW$ (see \eqref{eqn:lin_pca}), contrastive models trained with \ref{eq:dot} based losses do not seem to be equivalent to PCA for a constant matrix $\bC_\vartheta$. 
\end{remark}

\subsection{Cosine similarity Objective is Close to PCA for Wide Networks}\label{sec:CV}

Due to the near constancy of NTK under \ref{eq:cos} based losses, we investigate if it is possible to relate \ref{eq:cos} based trained contrastive models with PCA of a fixed matrix, potentially $\bC_\vartheta = \widetilde\bC_{\bV}(0)$. A direct equivalence seems complicated due to the normalization terms of \ref{eq:cos}. However, we note that with orthogonality on $\bW$, the cosine similarity measure can be bounded from below as
\begin{align}
    \label{eqn:cos_approx}
    \frac{f^{\perp}(\bx;\btheta )^\top f^{\perp}({\tilde\bx;\btheta })}{(\norm{f^{\perp}(\bx;\btheta )} + \delta)(\norm{f^{\perp}({\tilde\bx;\btheta })} + \delta)} \geq \frac{ \left( {\bW^\top}\phi({\bV\bx}) \right)^\top \left( {\bW^\top}\phi({\bV\tilde\bx })\right)}
    {(\norm{\phi({\bV\bx})} + \delta')(\norm{\phi({\bV\tilde\bx })} + \delta')}
\end{align}
since $\norm{{\bW^\top}\phi({\bV\bx}) }_2 \leq \norm{{\bW^\top}}_2 \norm{\phi({\bV\bx})}_2$ and $ \norm{{\bW}}_2 = 1$ (where $\delta':=\sqrt{M}\delta$). The inequality \eqref{eqn:cos_approx} can be used to define a modified loss $\calL(\calD)$ where $s(\bx,\tilde\bx)$ is defined as the right hand side of \eqref{eqn:cos_approx}. Minimizing the corresponding loss \eqref{eqn:orth_w2} with \eqref{eq:orth_nn} can now be written as
\begin{align}
   &\text{\eqref{eq: PCA} with } \bC_\vartheta=\widetilde\bC_V=\frac{\bC_V + \bC_V^\top}{2};~%\nonumber\\ 
  \bC_V \text{$=$} \frac{1}{N} \sum_{n = 1}^N \sum_{q = 1}^Q  
\frac{\alpha_q\phi({\bV\bx_{n,q}})\phi({\bV\bx_n})^\top}{(\norm{\phi({\bV\bx_{n,q}})} + \delta')(\norm{\phi({\bV\bx_n})} + \delta')}  \label{eq:C_v} 
\end{align}
Note that \eqref{eq:C_v} is of the form \eqref{eq: PCA}, however, $\bC_V$ is still dependent on $\bV$ and hence the optimization in \eqref{eqn:orth_w2} with \eqref{eq:orth_nn} is performed over both $\bV$ and $\bW$. Lemma~\ref{lemm:cw_change} below shows that for wide networks trained with \ref{eq:cos} based losses, $\bC_V(0)$ is close to $\bC_V(t)$ in Frobenius norm. Hence $\widetilde\bC_V$ also remains almost constant, which intuitively suggests that \eqref{eq:C_v} becomes ``close" to \eqref{eq: PCA}. 
\begin{lemma}[Constancy of $C_V(t)$]
\label{lemm:cw_change}
    Under  Assumptions~\ref{assm:delta}--\ref{assm:activation} and constraint $\bW^{\top}\bW = \bbI_Z$, consider training $f^{\perp}(\cdot)$ in \eqref{eq:orth_nn} for $t$ iterations using Grassmannian gradient descent\footnote{In short, following \cite{edelman1998geometry}, the derivative of a function $g(\cdot)$ restricted to a Grassmannian manifold can be obtained by left-multiplying $1-g(\cdot) g(\cdot)^\top$ to the unrestricted derivative of $g(\cdot)$.}under losses of the form \eqref{eq:C_v} with learning rate $\eta$. Then 
    %\begin{align*}
    $ \norm{ \bC_V(t) - \bC_V(0) }_{F} \leq \kappa \frac{t}{\sqrt{M}}$,
    where $\kappa := 16{\delta^{-2}}\eta Q^2L_{\phi}^2c_{in}^2\sqrt{D}$.
    %    \end{align*}
\end{lemma}

\begin{figure}[t]
    \centering
    \includegraphics[width = 0.98\textwidth]{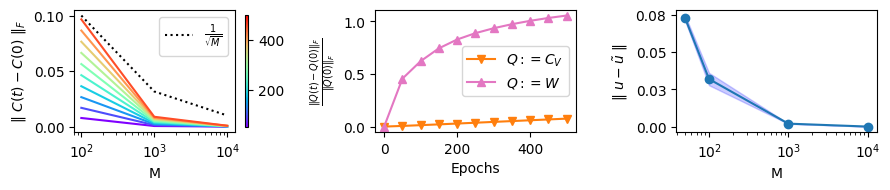}
    \caption{(left) Change in $\bC_V$ during training for varying depths $M$.  Time steps are indicated by color.
   (middle) Training rate of $\bC_V$ and $\bW$. Plotted are $\frac{\norm{\bC_V(t) - \bC_V(0)}_F}{\norm{ \bC_V(0)}_F}$ and $\frac{\norm{\bW(t) - \bW(0)}_F}{\norm{ \bW(0)}_F}$.
  (right) Difference in output when $\bC_V$ is frozen and when $\bC_V$ is trained for varying widths $M$.}
    \label{fig:PCA Theory}
\end{figure}

\paragraph{Numerical Simulation on MNIST.}  We train a network of the form \eqref{eq:orth_nn} using a loss of the form \eqref{eq:C_v}. We then examine the evolution of the quantity $\norm{ \bC_V(t) - \bC_V(0)  }_{F}$ with training across varying widths. The results are shown in Figure~\ref{fig:PCA Theory} (left), where colors indicate the epochs $(t\in[500])$. While the difference increases slightly with training, it goes down roughly as $\frac{1}{\sqrt{M}}$ with an increase in width, which is in line with the behaviour predicted in Lemma~\ref{lemm:cw_change}. In addition, we observe in Figure~\ref{fig:PCA Theory} (middle) that $\bW(t)$ changes significantly faster than $\bC_V(t)$ during training; this suggests that the $\bW$ that is learned is indeed the PCA of a $\widetilde\bC_V(t)$ that is close to $\widetilde\bC_V(0)$.

\subsection{Representations learned from PCA of $\widetilde C(0)$ and PCA of $\widetilde C(t)$ are close}\label{sec:representation}
We characterize the difference between the representations learned by performing  PCA on $\widetilde\bC_V(0)$ and on $\widetilde\bC_V(t)$. 
Lemmas~\ref{lemm:cw_change}--\ref{lemma:rep_close} suggest that training \eqref{eq:orth_nn} under \eqref{eqn:orth_w2} could be close to PCA on $\bC_V(0)$. 
\begin{lemma}[Perturbation bound on representation]
\label{lemma:rep_close}
Let $u(\textbf{x};\bW^*,\bC_\vartheta)$ be the representation obtained from \eqref{eq:orth_nn} with $\bW = \bW^*$, where $\bW^*$ is obtained by solving \eqref{eq: PCA}  for a fixed $\bC_\vartheta$. Under Assumptions~\ref{assm:delta}--\ref{assm:activation}, let $\bW^*,\widetilde\bW^*$ be the solutions of \eqref{eq: PCA} obtained through PCA on fixed $\widetilde\bC_V(0)$ and $\widetilde\bC_V(t)$ respectively. Let $\lambda_{Z},\lambda_{Z+1}$ be $Z^{\text{th}}$ and $(Z+1)^{\text{th}}$ eigenvalues of $\widetilde\bC_V(0)$. Let $\zeta = 4\delta^{-1}\eta Q\sqrt{D} L_{\phi}^2c_{in}^2$ and $\xi = 2^{\frac{7}{2}}\delta^{-1} QDc_{in}(L_{\phi}c_{\theta} + \abs{\phi(0)})$. There exists an orthogonal matrix $\bO$ such that 
\begin{align*}
\norm{\bO^\top u(\textbf{x};\widetilde\bW^*,\widetilde\bC_V(t)) - {u}(\textbf{x};\bW^*,\widetilde\bC_V(0))} \leq \zeta\frac{t}{\sqrt{M}}\left(1 + \frac{\xi \log M}{\lambda_{Z} - \lambda_{Z+1}} \right).
\end{align*}
\end{lemma}
%We now validate Lemma~\ref{lemma:rep_close} empirically and note that
%We outline the exact missing link in Remark~\ref{rem: open problem}. 

\paragraph{Numerical Simulation on MNIST.}
Consider training (\ref{eq:orth_nn}) according to (\ref{eq:C_v}) under two settings: with $\widetilde\bC_V(t)$ as a trainable matrix and with $\widetilde\bC_V(0)$ as a fixed matrix. We look at the fractional difference between the learned representations after training in the two settings. In Figure~\ref{fig:PCA Theory} (right), the mean fractional difference of learned representations computed across samples is plotted. As expected, the difference goes down to zero as width increases.

\section{Further discussion and open problems}\label{sec:discussion}
\paragraph{Effect of initialization on dimension collapse:}%\label{sec: initialization}
For the presented analysis, we only have a mild assumption on the initialization of weights (Assumption~\ref{assm:w_init}) that holds with high probability. While this assumption is sufficient for proving the constancy of NTK results in Section~\ref{sec: constancy}, additional assumptions may be needed to obtain \emph{meaningful} representations. While learning representations, it is desirable to avoid \emph{dimension collapse}. Dimension collapse, in the context of linear contrastive models, has been shown for \ref{eq:dot} \citep{esser2023representation}. Using NTK, we show that certain initialization schemes can cause dimension collapse even if \ref{eq:cos} based losses are used:
\begin{proposition}
    \label{prop:eq_we}
    Consider a neural network of the form (\ref{eqn: ntk_nn}) trained using a loss of the form (\ref{eqn:loss}) using gradient descent. Then for any input $\bx$,
    % \begin{align*} 
     $   {\bW_{.i}}(0) = {\bW_{.j}}(0)\Rightarrow f_i(\bx ; {\btheta(t)}) = f_j(\bx ; {\btheta(t)}),\  \forall \; t \geq 0.$
    % \end{align*}
\end{proposition}
 Thus, collapse occurs irrespective of whether the NTK remains constant or not, and hence, this is applicable irrespective of the width of the neural network. Further, the result holds for both dot product and cosine similarity based losses. The result also holds for analytical NTK $\bK^*$; if $\bK^*$ is used in the dynamics in Lemma~\ref{lemma:ntk_dyn}, then $    f_i(\bx ;\btheta(0)) = f_j(\bx ;\btheta(0))\Rightarrow f_i(\bx ;\btheta(t)) = f_j(\bx ;\btheta(t)),\  \forall t\geq0.$

\begin{figure}
    \centering
    \includegraphics[width = 0.98\textwidth]{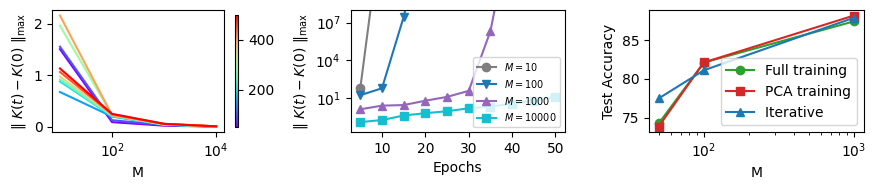}
    \caption{ 
    (left) Change in empirical NTK with \emph{cosine similarity} for 3  layer networks of varying widths (time steps indicated by color). 
    (middle) Maximum entry-wise change in empirical NTK with \emph{dot product similarity} for $3$  layer networks of varying widths.
  (right) Accuracy on downstream task for PCA on $\widetilde\bC(0)$, fully trained model and proposed iterative algorithm over $5$ updates. }
    \label{fig:additional_experiments}
\end{figure}

\paragraph{Empirical observations beyond the theory:}
\textbf{(i)} \emph{Deep networks:}
While Theorem~\ref{thm:ntk_conv} has been shown to hold only in the case of neural networks with a single hidden layer, we expect it to hold for deep networks as well. We experimentally examine the case with $3$ hidden layers in Figure~\ref{fig:additional_experiments} (left). The results are similar to the case of a single hidden layer, as we again observe a decay with width that is roughly $\frac{1}{\sqrt{M}}$ up to scaling.
Along similar lines, we observe in Figure~\ref{fig:additional_experiments} (middle) that the NTK for three hidden layer networks optimized with \ref{eq:dot} based losses diverges, similar to the single hidden layer case.
\textbf{(ii)} \emph{Iterative learning of the trace maximization problem:} Combining the findings from Section~\ref{sec: PCA connection}, we propose an alternative optimization procedure to solve \eqref{eq: PCA} under $\bA:=\widetilde\bC_V(t)$. As $\bC_V$ updates slower than $\bW$ as shown in Figure~\ref{fig:PCA Theory} (middle), and since for a fixed $\widetilde\bC_V$ the optimal $\bW$ can be obtained using PCA, an alternative update could be by iteratively (a) updating $\bW$ by solving PCA on $\widetilde\bC_V(t-1)$ and (b) updating $\bV$ by running one step of gradient descent. We validate this approach in the following.
\textbf{(iii)} \emph{Predictive accuracy in downstream tasks:} Extending the results from  Section~\ref{sec:representation}, we show in Figure~\ref{fig:additional_experiments} (right) that the representations obtained with PCA on $\widetilde\bC_V(0)$ and fully training \eqref{eq:orth_nn} under \eqref{eqn:orth_w2} are close to each other with regards to downstream test accuracy for a simple linear classifier trained on the representations. While Section~\ref{sec:representation} provides results on the behaviour of the PCA for different time-steps of $\widetilde\bC_V$, Figure~\ref{fig:additional_experiments} (right) suggests that this similarity extends to PCA on $\widetilde\bC_V(0)$ and fully trained networks as well.
In addition, we also observe that the iterative optimization (as proposed above) performs well, especially for smaller widths (possibly due to the larger change in $\widetilde\bC_V$  for small $M$).

\paragraph{Open problem (Missing link in the claim contrastive models perform PCA):}
%\label{rem: open problem}
In Lemma~\ref{lemma:rep_close}, we compare the solutions obtained by PCA on $\widetilde\bC_V(0)$ and $\widetilde\bC_V(t)$. But in Figure~\ref{fig:PCA Theory} (middle), we observe that  $\bC_V$ and $\bW$ evolve simultaneously, even though at different rates. The open problem then, is the exact connection between \emph{fully trained contrastive models and the PCA solution.} We believe that a direct analysis, such as the one considered for Theorem~\ref{thm:ntk_conv}, may not work. This is because the solutions might differ significantly at convergence even for closely initialized models if the eigengap is not significant. Therefore, a spectral viewpoint combined with the analysis of gradient descent steps is necessary to bound the deviation.

\paragraph{Open problem (NTK and PCA at convergence):}
Theorem~\ref{thm:ntk_conv} shows that, for wide networks, the NTK remains nearly constant for $M^{1/6}$ time-steps (in fact, one can also show constancy till $O(M^\alpha)$ steps for $\alpha<\frac14$). While such results are in line with initial NTK analysis in the supervised setting (e.g. \cite{jacot2018neural}), \emph{it is an open question whether the constancy of NTK holds until convergence}. 
In fact, Figure \ref{fig:1layer} suggests that for wide networks, NTK remains constant beyond $M^\alpha$ steps.
A potential approach to prove constancy till convergence would be to investigate the stationary points associated with cosine similarity based losses and verify if they are attained within $M^\alpha$ steps. While this is a crucial open question, we believe that the presented results provide the first valuable insights into the constancy of NTK under contrastive losses, beyond the squared error.

In  Lemma~\ref{lemma:rep_close}, we show that the outputs of the two considered trace maximization problems are close for $O({t} (\log M)/{\sqrt{M}})$ steps. As the expression is in terms of time-steps, \emph{the question remains if they are still close at convergence}. While we do not have a precise characterization of such results, a possible approach could be to extend \cite{xu2021comprehensively}, who show that for $Z=1$, PCA converges in roughly $O(\log(M))$ steps under Riemannian gradient descent.

% Acknowledgments---Will not appear in anonymized version
\acks{The work was done when G. G. Anil was at TU Munich, supported by the German Academic Exchange Service (DAAD) through the DAAD KOSPIE fellowship, 2023. This work is also supported by the German Research Foundation (DFG) through the Priority Program SPP 2298 (project GH 257/2-1).}
\bibliography{coltbib}

\clearpage
\appendix

\section*{Appendix}

In the appendix we provide the following additional material:
\begin{enumerate}
    % \item[\ref{app: related work}] Additional related work.
    \item[\ref{app: proofs}] Proofs for results in the main paper.
    \item[\ref{app: additioanl plots}] Additional plots and experimental results.
\end{enumerate}

% \clearpage
% \section{Related Work}\label{app: related work}

% In practice a plethora of SSL strategies have been proposed and evaluated over the past years across multiple domains; many of them demonstrating excellent performance empirically on several benchmarks \citep{hjelm2018learning, wu2018unsupervised, bachman2019learning, devlin2018bert, misra2020self, henaff2020data, radford2021learning, caron2021emerging, assran2022masked, wang2023masked}. 

% In recent works the main focus has firstly been on modeling generalization error for downstream tasks on embeddings obtained by SSL \citep{Arora2019ATA,GeTFJ-2303-01566,BansalKB21,LeeLSZ21,ChenZXCD0TZC22,SaunshiMA21,ToshK021,WeiXM21,0002NN22}, and secondly on analysing the spectral and isoperimetric properties of data augmentation \citep{BalestrieroL22,HanYZ23,Zhuo0M023}.

% \clearpage
\section{Proofs}\label{app: proofs}
\localtableofcontents

\subsection{Lemma \ref{lemma:hessian} [Bound on the norm of the Hessian]}
     Under Assumptions~\ref{assm:data}, \ref{assm:activation}, \ref{assm:gradient}, consider the neural network defined in \eqref{eqn: ntk_nn}. If the change in weights during training is bounded as 
     \begin{align*}
         \norm{\bW(t) - \bW(0)}_F + \norm{\bV(t) - \bV(0)}_F \leq R,
         % \label{eq: R bound}
         \end{align*}
          then, $\forall \; z \in [Z]$, with $\alpha_1 = 4\beta_{\phi} c_{in}^2 L_{\phi}$ and $\alpha_2 = 4L_{\phi}c_{in}(1 + \beta_{\phi} c_{in}s_0 c_s)$,  the $z^\text{th}$ Hessian is bounded as:
   $  \norm{\bH^{(z)}(\bx ; \btheta(t))}_2 \leq \frac{\alpha_1 R + \alpha_2}{\sqrt{M}}.$
\begin{proof}
   % Consider the $z^{\text{th}}$ Hessian $\bH^{(z)}(\bx ; \btheta)$. Let us denote the vector containing all entries of $\bV$ as $\theta_V$ and the vector containing all the entries of $z^{\text{th}}$ column of $\bW$ as $\theta_{W_z}$. After appropriate permutations of rows and columns, we can write:
   % \begin{align*}
   %     \bH^{(z)}(\bx ; \btheta) = \begin{bmatrix}
   %     \frac{\partial^2f_z}{\partial^2 (\theta_V)^0} & 2 &3\\
   %     4&5&6
   %     \end{bmatrix}
   % \end{align*}
   Note that the $z^{\text{th}}$ entry of the embedding $f(\cdot)$ can be written as
\begin{align*}
    f_z(\bx; \btheta) = \frac{1}{\sqrt{M}}(\bW^\top)_{z.} \phi(\bV\bx).
\end{align*}
Therefore, each $f_z(\bx; \btheta)$ can be though of as a single-dimensional output of a neural network with weights $\{\bV,\bW_{z.} \}$. Further,
\begin{align*}
    &\norm{\bV(t) - \bV(0)}_F + \norm{\bW(t) - \bW(0)}_F \leq R\\
    \implies& \norm{\bV(t) - \bV(0)}_F + \norm{\bW^\top _{z.}(t) - \bW^\top _{z.}(0)}_F \leq R
\end{align*}
since $ \norm{\bW^\top _{z.}(t) -\bW^\top _{z.}(0)}_F \leq \norm{\bW(t) - \bW(0)}_F$. Let the Hessian corresponding to this neural network be denoted as $\Tilde{\bH}^{(z)}(\bx ; \btheta(t))$. Note that $\Tilde{\bH}^{(z)}(\bx ; \btheta(t))$ satisfies all conditions required (linear final layer and Gaussian initialization) by Theorem~7.1 of \cite{liu2020toward}, and hence we can use this theorem to bound $\norm{\Tilde{\bH}^{(z)}(\bx ; \btheta(t))}_2$. To connect this to $\norm{\bH^{(z)}(\bx ; \btheta(t))}_2$, note that after appropriate permutations of rows and columns (note that spectral norm is invariant to permutations), we can write:
\begin{align*}
    \bH^{(z)}(\bx ; \btheta(t)) &= \begin{bmatrix}
    \Tilde{\bH}^{(z)}(\bx ; \btheta(t)) & \mathbf{0} \\
    \mathbf{0} & \mathbf{0}
    \end{bmatrix}
\end{align*}
and therefore  $\norm{\bH^{(z)}(\bx ; \btheta(t))}_2 = \norm{\Tilde{\bH}^{(z)}(\bx ; \btheta(t))}_2$. Hence, we can apply Theorem~7.1 of \cite{liu2020toward} to $\norm{\bH^{(z)}(\bx ; \btheta(t))}_2$. Now, from the proof of Theorem~7.1 in \cite{liu2020toward}, we have:
\begin{align*}
    \norm{H^{(z)}(\bx; \btheta(t))}_2 \leq \frac{L^2 C'(R)}{\sqrt{M}}
\end{align*}
where $L$ is the number of layers in the neural network and 
\begin{align*}
    C'(R) &= \beta_{\phi} c_{in}^2\sum_{l' = 1}^{L-1} L_{\phi}^{2l'-1}(c_0 + R)^{2l'-2}C_{\bb}^{(l')}(R) + L_{\phi}^{L-1}(c_0 + R)^{L-2}c_{in}
\end{align*}
For $L = 2$, which is the case we are interested in, this reduces to:
\begin{align*}
    C'(R) &= \beta_{\phi} c_{in}^2 L_{\phi}C_{\bb}^{(1)}(R) + L_{\phi}c_{in} 
\end{align*}
In our setup, we have $c_0 = c_s$, where $c_s$ is defined in Assumption~\ref{assm:gradient} and $C_{\bb}^{(1)}(R) = s_0c_0 + R$, where $s_0$ is defined in Assumption~\ref{assm:gradient}.
So for each Hessian matrix, we have:
\begin{align}
    \label{eq: spec_bound}
    \norm{H^{(z)}(\bx; \btheta(t))}_2 \leq \frac{\alpha_1R + \alpha_2}{\sqrt{M}}
\end{align}
where $\alpha_1 = 4\beta_{\phi} c_{in}^2 L_{\phi}$ and $\alpha_2 = 4L_{\phi}c_{in}(1 + \beta_{\phi} c_{in}s_0 c_s)$. Note that in general, for $L$ layers, the spectral norm has a bound of the form $ \norm{H^{(z)}(\bx; \btheta(t))}_2 \leq O(\frac{R^{3L}}{\sqrt{M}})$. However, for the special case of $L = 2$, it is possible to further reduce it to the form in (\ref{eq: spec_bound}). 
\end{proof}

\subsection{Lemma \ref{lemma:ntkhess}  [Bound on the change in NTK]}

    Define $\bbS := \{\bs \in \mathbb{R}^p; \norm{\bs - \bs(0)} \leq R\}$, where $p$ is the total number of learnable parameters in \eqref{eqn: ntk_nn}. Assume that for any input $\bx$, $\norm{\bH^{(z)}(\bx; \bs)}_2 \leq \epsilon$ and $\norm{\nabla_{\bs}f_z({\bx; {\bs}})}_2 \leq c_0$, $\forall \; z \in [Z]$ and $\forall \; \bs \in \bbS$. Then, for any inputs $\bx, \tilde\bx$, $\forall \; \bs \in \bbS$ and $\forall \; i,j \in [Z]$, $\abs{\bK_{ij}(\bx , {\tilde\bx }; {\bs}) - \bK_{ij}(\bx , {\tilde\bx }; {\bs(0)})} \leq 2\epsilon c_0 R$.
\begin{proof}
We start by writing out and expanding the difference in the $(i,j)^{\text{th}}$ kernel entry when there is a change in $\bs$ from $\bs(0)$ to $\bs$:
\begin{align*}
     &\abs{\bK_{ij}(\bx, \by; \bs ) - \bK_{ij}(\bx, \by; \bs(0)  )} \\
     &= \abs{ {\frac{\partial f_i(\bx, \bs )}{\partial \bs }}^\top \frac{\partial f_j(\by, \bs )}{\partial \bs } - {\frac{\partial f_i(\bx, \bs(0)  )}{\partial \bs(0)  }}^\top \frac{\partial f_j(\by, \bs(0)  )}{\partial \bs(0)  }} \\
     &= \abs{ \left( {\frac{\partial f_i(\bx, \bs )}{\partial \bs }} 
 - {\frac{\partial f_i(\bx, \bs(0)  )}{\partial \bs(0)  }} \right)^\top \frac{\partial f_j(\by, \bs )}{\partial \bs } + {\frac{\partial f_i(\bx, \bs(0)  )}{\partial \bs(0)  }}^\top  \left( \frac{\partial f_j(\by, \bs )}{\partial \bs } - \frac{\partial f_j(\by, \bs(0)  )}{\partial \bs(0)  } \right)} \\
 &\leq \norm{{\frac{\partial f_i(\bx, \bs )}{\partial \bs }} 
 - {\frac{\partial f_i(\bx, \bs(0)  )}{\partial \bs(0)  }}} \norm{\frac{\partial f_j(\by, \bs )}{\partial \bs }} + \norm{{\frac{\partial f_i(\bx, \bs(0)  )}{\partial \bs(0)  }}}\norm{\frac{\partial f_j(\by, \bs )}{\partial \bs } - \frac{\partial f_j(\by, \bs(0)  )}{\partial \bs(0)  }}
\end{align*}
where we have used Triangle and Cauchy-Schwartz inequalities. Using the fact that $\norm{\nabla_{\bs} f_z(\bx; \bs )}$, $ \norm{\nabla_{\bs} f_z(\by; \bs )} \leq c_0$, $ \; \forall \; \bs  \in S, z \in [Z]$ (from the lemma assumptions), we can simplify the expression to:
\begin{align*}
    &\abs{\bK_{ij}(\bx, \by; \bs ) - \bK_{ij}(\bx, \by; \bs(0)  )} \\
    &\leq c_0 \left( \norm{{\frac{\partial f_i(\bx, \bs )}{\partial \bs }} - {\frac{\partial f_i(\bx, \bs(0)  )}{\partial \bs(0)  }}} +  \norm{\frac{\partial f_j(\by, \bs )}{\partial \bs } - \frac{\partial f_j(\by, \bs(0)  )}{\partial \bs(0)  }} \right).
\end{align*}
From Proposition~2.3 of \cite{liu2020linearity}, we know that the remaining terms can be bounded in terms of the Hessian and $R$ as follows:
\begin{align*}
    \norm{{\frac{\partial f_i(\bx, \bs )}{\partial \bs }} - {\frac{\partial f_i(\bx, \bs(0)  )}{\partial \bs(0)  }}} &\leq \max_{\bs  \in S}\norm{H_i(\bx; \bs )} R, \\
    \norm{{\frac{\partial f_j(\by, \bs )}{\partial \bs }} - {\frac{\partial f_j(\by, \bs(0)  )}{\partial \bs(0)  }}} &\leq \max_{\bs  \in S}\norm{H_j(\by; \bs )} R.
\end{align*}
Combining the above results, we can bound the kernel difference as:
\begin{align*}
    \abs{\bK_{ij}(\bx, \by; \bs ) - \bK_{ij}(\bx, \by; \bs(0)  )} &\leq  2\epsilon c_0 R
\end{align*}
which concludes the proof.
\end{proof}

\subsection{Lemma~\ref{lemma:ntk_dyn} [Contrastive learning dynamics in terms of NTK]}
 
    Consider training a neural network of the form (\ref{eqn: ntk_nn}) using a loss $l(\cdot)$ of the form (\ref{eqn:loss}) under gradient flow on dataset $\mathcal{D}$. Let  $g_i(\bx, \tilde\bx;\btheta(t)  ) := \frac{\partial s(\bx, \tilde\bx  )}{\partial f_i(\bx;\btheta(t))}$. Then, for $z \in [Z]$, the representation of an arbitrary input $\tilde\bx $ evolves as:
    \begin{align*}
    \frac{\partial {f}_z({\tilde\bx;\btheta(t) })}{\partial t}=
    -\frac{1}{N}\sum_{n, q} \frac{\partial l(\cdot)}{\partial s(\bx_n, \xpair )}
         &\left[ \sum_{i=1}^{Z} \left[\bK_{zi}(\tilde\bx  , \bx_n; \btheta(t))g_i(\bx_n,\xpair;\btheta(t) )  \right.\right.\\
         &\left.\qquad\qquad+ \bK_{zi}(\tilde\bx  , \xpair ; \btheta(t))g_i(\xpair , \bx_n;\btheta(t)) \right]\big].
    \end{align*}
\begin{proof}
We are interested in computing the dynamics of $f_z(\tilde\bx;\btheta )$. Note that the dynamics can be expressed as
\begin{align}\label{eq:app:dynamics}
    \dot{f}_z(\tilde\bx) = \left( \frac{d f_z(\tilde\bx;\btheta)}{d \btheta} \right)^\top  \frac{d \btheta}{dt}.
\end{align}
Since we are optimizing using gradient flow, $\frac{d \btheta}{dt} = - \nabla_{\btheta}\mathcal{L}$. We now compute $\nabla_{\btheta}\mathcal{L}$ as follows:
\begin{align*}
    \nabla_{\btheta}\mathcal{L} &= \frac{1}{N} \sum_{n} \nabla_{\btheta} l\left( \{s(\bx_n, \xpair )\}_{q = 1}^{Q}\right) \\
    &= \frac{1}{N} \sum_{n, q} \frac{\partial l}{\partial s(\bx_n , \xpair )} \nabla_{\btheta} s(\bx_n , \xpair ) \\
    &= \frac{1}{N} \sum_{n, q} \frac{\partial l}{\partial s(\bx_n , \xpair )} \left[ \sum_{i = 1}^Z \left[ \frac{\partial s(\bx_n , \xpair )}{\partial f_i(\bx_n;\btheta )} \cdot \frac{\partial f_i(\bx_n;\btheta )}{\partial \btheta} + \frac{\partial s(\bx_n , \xpair )}{\partial f_i(\xpair;\btheta )} \cdot \frac{\partial f_i(\xpair;\btheta )}{\partial \btheta}  \right] \right]  .
\end{align*}
Therefore using $ \nabla_{\btheta}\mathcal{L} $ in \eqref{eq:app:dynamics} we obtain the dynamics as
\begin{align*}
    \dot{f}_z(\tilde\bx) &=  \frac{- 1}{N} \sum_{n, q} \frac{\partial l}{\partial s(\bx_n , \xpair )} \left[ \sum_{i = 1}^Z  \left( \frac{d f_z(\tilde\bx;\btheta)}{d \btheta} \right)^\top  \left[ \frac{\partial s(\bx_n , \xpair )}{\partial f_i(\bx_n;\btheta )} \cdot \frac{\partial f_i(\bx_n;\btheta )}{\partial \btheta} \right.\right.\\
    & \qquad \left.\left.+ \frac{\partial s(\bx_n , \xpair )}{\partial f_i(\xpair ;\btheta)} \cdot \frac{\partial f_i(\xpair ;\btheta)}{\partial\btheta}  \right] \right]  \\
    &= \frac{- 1}{N} \sum_{n, q} \frac{\partial l}{\partial s(\bx_n , \xpair )} \left[ \sum_{i = 1}^Z  \left[ \frac{\partial s(\bx_n , \xpair )}{\partial f_i(\bx_n ;\btheta)} \cdot \bK_{z i}(\tilde\bx, \bx_n;\btheta )\right.\right. \\
   & \qquad\left.\left.+ \frac{\partial s(\bx_n , \xpair )}{\partial f_i(\xpair;\btheta )} \cdot \bK_{z i}(\tilde\bx, \xpair ;\btheta)  \right] \right]
\end{align*}
which, after substituting $g_i(\textbf{x}, \tilde\bx;\btheta(t)  ) := \frac{\partial s(\textbf{x}, \tilde\bx  )}{\partial f_i(\bx;\btheta(t))}$, provides the expression in Lemma~\ref{lemma:ntk_dyn} and thus concludes the proof.
\end{proof}

\subsection{Proposition \ref{prop:lin_change} [NTK under dot product does not remain constant]}

    % % \label{prop:lin_change}
    % % Under  Assumptions~\ref{assm:}~-~\ref{assm:} \note{add}  and
    % For $D=Z=1$, linear loss $(l(a):=a)$, \ref{eq:dot} similarity and triplet setting $(\calD = \{x_n,x_n^+,x_n^-\}_{n=1}^N)$ in \eqref{eqn:loss}, the optimization is:
    % % \begin{align*}
    % $
    %      \min_\btheta   \frac{1}{N}\sum_{n = 1}^N  f(x_n;\btheta)\left(f(x_n^-;\btheta) - f(x_n^+;\btheta)\right)$.
    % % \end{align*}
    % % . Assume a linear network $(\phi(a):=a)$ and linear loss $(l(a):=a)$, trained under gradient flow. for a triplet setting $\calD = \{x_n,x_n^+,x_n^-\}_{n=1}^N$ can be written as 
    % % Assume that ${\bw}^\top {\bv}$ is non-zero at initialization. Consider $ f_{lin}(\cdot)$, trained using the loss \eqref{eqn:lin_loss} (which is a \ref{eq:dot} loss) using gradient flow. 
    % Consider a network \eqref{eqn: ntk_nn} with linear activation $(\phi(a):=a)$, weights initialised as independent $\mathcal{N}(0,1)$, and trained via gradient flow.
    % \\
    % There is a dataset such that, with probability at least  $1 - \frac{25}{\sqrt{M}}$, for a time step $\Tilde{t} \in (0, \log M)$ and any input pair $x,\Tilde{x}$ with $x\Tilde{x} \neq 0$, the NTK satisfies 
    %   $    \abs{\bK(x, \tilde x; \btheta(t)) -  \bK(x, \tilde x; \btheta(0))} \rightarrow \infty$ as $t \rightarrow \Tilde{t}$.
    %     % \end{align*}

   For $D=Z=1$, linear loss $(l(a):=a)$, \ref{eq:dot} similarity and triplet setting $(\calD = \{x_n,x_n^+,$ $x_n^-\}_{n=1}^N)$ in \eqref{eqn:loss}, the optimization is:
    % \begin{align*}
    $
         \min_\btheta   \frac{1}{N}\sum_{n = 1}^N  f(x_n;\btheta)\left(f(x_n^-;\btheta) - f(x_n^+;\btheta)\right)$.
    % \end{align*}
    % . Assume a linear network $(\phi(a):=a)$ and linear loss $(l(a):=a)$, trained under gradient flow. for a triplet setting $\calD = \{x_n,x_n^+,x_n^-\}_{n=1}^N$ can be written as 
    % Assume that ${\bw}^\top {\bv}$ is non-zero at initialization. Consider $ f_{lin}(\cdot)$, trained using the loss \eqref{eqn:lin_loss} (which is a \ref{eq:dot} loss) using gradient flow. 
    Consider a network \eqref{eqn: ntk_nn} with linear activation $(\phi(a):=a)$, weights initialised as independent $\mathcal{N}(0,1)$, and trained via gradient flow.
    \\
    There is a dataset such that, with probability at least  $1 - \frac{25}{\sqrt{M}}$, for a time step $\Tilde{t} \in (0, \log M)$ and any input pair $x,\Tilde{x}$ with $x\Tilde{x} \neq 0$, the NTK satisfies 
      $    \abs{\bK(x, \tilde x; \btheta(t)) -  \bK(x, \tilde x; \btheta(0))} \rightarrow \infty$ as $t \rightarrow \Tilde{t}$.
        % \end{align*}
\begin{proof}    
Let us first recall the definition of a linear neural network with one dimensional input $\bx\in\bbR$ and one dimensional output $(Z=1)$:
\begin{align*}
     f(x;\btheta) = \frac{1}{\sqrt{M}} \sum_{m = 1}^M \bw_m \bv_m  x .
\end{align*}
Plugging this into the loss function, we obtain:
\begin{align*}
\calL &= \frac{1}{N}\sum_{n = 1}^N  f(x_n;\btheta)\left(f(x_n^-;\btheta) - f(x_n^+;\btheta)\right)\\
     &= \frac{1}{N} \sum_{n = 1}^N \frac{1}{\sqrt{M}} \sum_{m = 1}^M \bw_m \bv_m  x_n \left(\frac{1}{\sqrt{M}} \sum_{m = 1}^M \bw_m \bv_m  \bx_n^- - \frac{1}{\sqrt{M}} \sum_{m = 1}^M \bw_m \bv_m  x_n^+  \right) \\
    &= \frac{-1}{{M}} \left( \sum_{m = 1}^M \bw_m \bv_m   \right)^2 \left (\frac{1}{N}  \sum_{n = 1}^N  x_n(x_n^+ - x_n^-)  \right) \\
    &= \frac{-C}{{M}} \left( \sum_{m = 1}^M \bw_m \bv_m   \right)^2
\end{align*}
where $C = \frac{1}{N}  \sum_{n = 1}^N  x_n(x_n^+ - x_n^-)$ is a data dependent constant. Recall that we aim to find a setting under which the NTK does not remain constant and as such it is sufficient to construct an example where this is the case. We therefore choose a dataset such that $C > 0$ (we do not have to explicitly construct this dataset for the proof, we just need the existence of such a dataset). 
Let us define the following for ease of notation:
\begin{align*}
    S &= \frac{1}{M} \left( \sum_{m = 1}^M \bw_m  \bv_m   \right),\\
    P &= \frac{1}{M} \left( \sum_{m = 1}^M (\bv_m^2 + \bw_m^2)  \right).
\end{align*}
% In this setting under gradient flow, we have:
We start by noting that we can write the NTK as:
\begin{align*}
    \bK(x, y; \btheta(t)) &= \frac{1}{M}\left( \sum_{m = 1}^M (\bv_m^2 + \bw_m^2)  \right)xy\\
    &= P(t)\ xy.
\end{align*}
Therefore as $xy$ is a fixed constant, to compute the change in NTK, it is sufficient to get an expression for the change of $P$ with respect to $t$, which we can write as:
\begin{align}\label{eq:app:dPdt}
    \frac{d P}{d t} &= \frac{2}{M} \left( \sum_{m = 1}^M \left({\bv_m }\frac{d \bv_m }{dt} + {\bw_m }\frac{d \bw_m }{dt}\right)  \right),
\end{align}
Now, we have, from the gradient flow assumption:
\begin{align}
    \label{eqn:w1_ev} \frac{d \bv_m }{d t} &= - \frac{d \mathcal{L}}{d \bv_m } = \frac{2C}{{M}} \left( \sum_{m' = 1}^M \bw_{m'} \bv_{m'}  \right) \bw_m = 2CS\bw_m,\\
    \label{eqn:w2_ev} \frac{d \bw_m}{d t} &= - \frac{d \mathcal{L}}{d \bw_m} = \frac{2C}{{M}} \left( \sum_{m' = 1}^M \bw_{m'} \bv_{m'}  \right) \bv_m  = 2CS\bv_m .
\end{align}
Using \eqref{eqn:w1_ev} and \eqref{eqn:w2_ev} in \eqref{eq:app:dPdt}, we can further rewrite $\frac{d P}{d t}$ in terms of $S$:
\begin{align}
    \frac{d P}{d t} &= \frac{2C}{{M}} \left( \sum_{m' = 1}^M \bw_{m'} \bv_{m'}  \right) \left[\frac{2}{M} \sum_{m = 1}^M ({\bv_m }{\bw_m } + {\bw_m }{\bv_m }) \right] \nonumber \\
   \label{eqn:p_dyn} &= 8C S^2 .
\end{align}
Now computing the change of $S$ w.r.t. $t$, we obtain:
\begin{align}
    \frac{d S}{d t} &= \frac{1}{M}  \left( \sum_{m = 1}^M \left(\bw_m   \frac{d \bv_m }{dt} + \bv_m   \frac{d \bw_m }{dt} \right)  \right) \nonumber\\
    &= \frac{2C}{{M}} \left( \sum_{m' = 1}^M \bw_{m'} \bv_{m'}  \right) \left[\frac{1}{M} \sum_{m = 1}^M ({\bv_m^2 } + {\bw_m^2 }) \right] \nonumber\\
    \label{eqn:s_dyn} &= 2C S P
\end{align}
From (\ref{eqn:p_dyn}) and (\ref{eqn:s_dyn}), we have:
\begin{align}
    \frac{dP}{dS} &= \frac{4S}{P} \nonumber \\
    \label{eqn:sp_rel} \implies S^2(t) &= \frac{P^2(t)}{4} - \alpha
\end{align}
where $\alpha = \frac{P^2(0)}{4} - S^2(0) $. Also, note that:
\begin{align}
    \label{eqn:p_pos} P^2(t) - 4\alpha \geq 0
\end{align}
for all $t$ since $S^2(t) \geq 0$. From (\ref{eqn:p_dyn}) and (\ref{eqn:sp_rel}), we have:
\begin{align}
    \frac{dP}{dt} &= 2C(P^2 - 4\alpha) \nonumber\\
    \label{eqn:p_time} \frac{dP}{P^2 - 4\alpha} &= 2Cdt
\end{align}
Now,
\begin{align*}
    4\alpha &= P^2 - 4S^2 \\
    &= \frac{1}{M^2}\left[ \left(\sum_{m = 1}^M (\bv_m^2  + \bw_m^2 ) \right)^2 - 4\left( \sum_{m = 1}^M \bv_m \bw_m \right)^2 \right] \\
    &= \frac{1}{M^2}\left[ \left(\sum_{m = 1}^M (\bv_m^2  + \bw_m^2  + 2\bv_m \bw_m)  \right) \left(\sum_{m = 1}^M (\bv_m^2  + \bw_m^2  - 2\bv_m \bw_m)  \right)  \right] \\
    &= \frac{1}{M^2} \left[\left(\sum_{m = 1}^M (\bv_m  + \bw_m )^2 \right) \left(\sum_{m = 1}^M (\bv_m  - \bw_m )^2 \right)  \right] \\
    &\geq 0
\end{align*}
Therefore, $\frac{dP}{P^2 - 4\alpha}$ is of the form $\frac{dx}{x^2 - a^2}$. (\ref{eqn:p_time}) can then be integrated using partial fractions method to obtain:
\begin{align*}
    {\abs{\frac{P - 2\sqrt{\alpha}}{P + 2\sqrt{\alpha}}}} = \beta e^{8C\sqrt{\alpha}t} 
\end{align*}
where $\beta = \abs{\frac{P(0) - 2\sqrt{\alpha}}{P(0) + 2\sqrt{\alpha}}}$. Now, using (\ref{eqn:p_pos}) and the fact that $P, 4\alpha \geq 0$, we have $P - 2\sqrt{\alpha} \geq 0$. Therefore, we have:
\begin{align*}
    {\frac{P - 2\sqrt{\alpha}}{P + 2\sqrt{\alpha}}} &= \beta e^{8C\sqrt{\alpha}t} \\
    \implies P(t) &= 2\sqrt{\alpha}\left( \frac{1 + \beta e^{8C\sqrt{\alpha} t}}{1 - \beta e^{8C\sqrt{\alpha} t}} \right)
\end{align*}
Note that since each $\bv_m, \bw_m$ is initialized as $\mathcal{N}(0, 1)$, $\bw^\top \bv \neq 0$ with probability $1$ at initialization. Hence, $S^2(0) > 0$ with probability $1$. Therefore, $P^2(0) - 4\alpha > 0$ and hence $P(0) - 2\sqrt{\alpha} > 0$. Consequently, $\beta > 0$ with probability 1. Let $\Tilde{t}$ indicate the time $t$ at which $1 - \beta e^{8C\sqrt{\alpha} t} = 0$. Then, $\Tilde{t} = \frac{1}{8C\sqrt{\alpha}}\log{\frac{1}{\beta}}$. Since $0 < \beta < 1 $, $\Tilde{t} > 0$. Clearly, as $t \rightarrow \Tilde{t}$, $P(t) \rightarrow \infty$. Since $\abs{\bK(x, y; \btheta(t)) - \bK(x, y; \btheta(0))}  = \abs{(P(t) - P(0))xy}$,
\begin{align*}
    \abs{\bK(x, y; \btheta(t)) - \bK(x, y; \btheta(0))} \rightarrow \infty
\end{align*}
if $xy\neq 0$.

\noindent In the final part of the proof, we show that $\Tilde{t} \in (0, \log M)$ with probability at least $1-\frac{25}{\sqrt{M}}$. We first compute a few  moments for the random variables $S(0)$ and $P(0)$.
Due to standard normal initialization of the weights, one can compute the following: 
$\mathbb{E}[P(0)] = 2, Var(P(0)) = \frac{4}{M}, \mathbb{E}[S(0)] = 0, \mathbb{E}[S(0)^2] = Var(S(0)) = \frac1M$.
%, \mathbb{E}[S(0)^4] = \frac{3}{M^2} + \frac{6}{M^3}$, 
and $Var(S(0)^2) = \frac{2}{M^2} + \frac{6}{M^3}$.
By applying Chebyshev's inequality, it immediately follows that
\begin{align*}
\mathbb{P}\left(|P(0)-2| > \frac12\right) \leq 4\cdot Var(P(0)) = \frac{16}{M}; \qquad
\mathbb{P}\left(|S(0)| > \frac12\right) \leq 4\cdot Var(S(0)) = \frac{4}{M}. 
\end{align*}
We further need to show that $S(0)^2 > \frac{1}{M^2}$ with high probability, but Chebyshev's inequality does not suffice to prove this (as $Var(S(0)^2)$ is too large).
To this end, we use Berry-Esseen theorem \cite[see, for instance,][Corollary 1]{KorolevShevtsova10}: Let $Z_1,\ldots,Z_M$ be i.i.d. variables with $\mathbb{E}[Z_m] = 0, \mathbb{E}[Z_m^2] = \sigma^2$ and $\mathbb{E}[|Z_m|^3] \leq \rho < \infty$. The deviation between cumulative distribution, $F_Y(\cdot)$, of $Y = \frac{1}{\sigma\sqrt{M}}\sum_m Z_m$ and standard normal distribution, $\Phi(\cdot)$, is bounded as
\[
\sup_x \left| F_Y(x) - \Phi(x) \right| < \frac{0.52\rho}{\sigma^3\sqrt{M}}.
\]
In the present context, $Z_m = w_mv_m$ satisfies $\mathbb{E}[Z_m] = 0, \mathbb{E}[Z_m^2] = \mathbb{E}[w_m^2]\mathbb{E}[v_m^2] = 1$ and $\mathbb{E}[|Z_m|^3] = (\mathbb{E}[|w_m|^3])^2 \leq \mathbb{E}[w_m^2]\mathbb{E}[w_m^4]=3$ (Cauchy-Schwartz inequality). Further, $Y = \sqrt{M}S(0)$.
Hence, we can bound:
\begin{align*}
    \mathbb{P}\left(S(0)^2 \leq \frac{1}{M^2}\right)
    = \mathbb{P}\left(|Y| \leq \frac{1}{\sqrt{M}}\right)
    &= F_Y\left(\frac{1}{\sqrt{M}}\right) - F_Y\left(\frac{-1}{\sqrt{M}}\right)
    \\&\leq \Phi\left(\frac{1}{\sqrt{M}}\right) - \Phi\left(-\frac{1}{\sqrt{M}}\right) + 2 \sup_x \left| F_Y(x) - \Phi(x) \right|
    \\&< \frac{2}{\sqrt{M}}\cdot\frac{1}{\sqrt{2\pi}} \; + \; 2\cdot \frac{1.56}{\sqrt{M}} < \frac{5}{\sqrt{M}}. 
\end{align*}
Combining all probability bounds, we have that, with probability at least $1-\frac{20}{M}-\frac{5}{\sqrt{M}} \geq 1 - \frac{25}{\sqrt{M}}$, it holds that $P(0) \in [1.5, 2.5], S(0) \in [-0.5,0.5]$ and $S(0)^2 > \frac{1}{M^2}$.
Under this condition, it immediately follows that $\alpha = \frac{P(0)^2}{4} - S(0)^2$ is bounded as $\frac14 < \alpha < 2$, while
\[\beta = \frac{P(0) - 2\sqrt{\alpha}}{P(0) + 2\sqrt{\alpha}} = \frac{4S(0)^2}{(P(0) + 2\sqrt{\alpha})^2} > \frac{4S(0)^2}{36} > \frac{1}{9M^2}.\]
Using the lower bounds of $\alpha,\beta$ in the expression for $\Tilde{t}$, it follows that
$\Tilde{t} < \frac{1}{2C}\log (3M)$ with probability at least $1 - \frac{25}{\sqrt{M}}$. 
One can choose the dataset such that $C>0$ is large enough to make the upper bound at most $\log M$.
\end{proof}

% &= \\
%     \implies \bv_m \frac{d \bw_m}{d t} &= \frac{2C}{{M}} \left( \sum_{m' = 1}^M \bw_{m'} \bv_{m'}  \right) (\bv_m )^2 \\
%     \implies \frac{d (\bv_m \bw_m)}{d t} &= \frac{2C}{{M}} \left( \sum_{m' = 1}^M \bw_{m'} \bv_{m'}  \right) (\bv_m )^2
% where we have used the fact that $\bv_m $ is not trained and hence is a constant. Summing up the equations over all $m$, we have:
% \begin{align*}
%     \frac{d \left( \sum_{m = 1}^M \bv_m \bw_m \right)}{d t} &= \frac{2C}{{M}} \left( \sum_{m' = 1}^M \bw_{m'} \bv_{m'}  \right) \left( \sum_{m = 1}^M (\bv_m )^2 \right)
% \end{align*}
% Define $S = \sum_{m = 1}^M \bv_m \bw_m$. Then we have:
% \begin{align*}
%     \frac{d S}{d t} &= \frac{2C}{{M}} \left( \sum_{m = 1}^M (\bv_m )^2 \right) S 
% \end{align*}
% Let us assume \bV is such that $\frac{1}{M}  \left( \sum_{m = 1}^M (\bv_m )^2 \right) = 1$. Note that this happens with high probability with the initialization under consideration. Then, we have:
% \begin{align*}
%     \frac{d S}{d t} &= 2C S
% \end{align*}
% Let us assume the dataset is such that $C > 0$. Then, on solving the differential equation, we have:
% \begin{align*}
%     S(t) = S(0)e^{2C t}
% \end{align*}
% Therefore:
% \begin{align*}
%     \frac{d \bw_m}{d t} &= \frac{2C \bv_m  S(0)}{M} e^{2C t} \\
%     \implies  \abs{\bw_m(t) - \bw_m(0)} &= \Omega \left(\frac{e^{2C t}}{M}   \right)
% \end{align*}
% Setting $\alpha = 2C$, the proof is complete.

%K_{z, i}(\by, \bx_n )
\subsection{Lemma~\ref{lemma:change} [Bound on element-wise change in weights under cosine similarity]}

    Under  Assumptions~\ref{assm:delta}~-~\ref{assm:activation}, consider losses of the form \eqref{eqn:loss} where \ref{eq:cos} is used. If a neural network $f(\cdot)$ as defined in \eqref{eqn: ntk_nn} is trained using gradient descent with learning rate $\eta$, at any time $t$, the change in weights are bounded as:
 $   \abs{\Delta \bV_{ij}(t)} \leq \frac{\beta_1}{\sqrt{M}} \; %\max_{ij} \left \{\abs{\bV_{ij}(t)} \right\}
    \maxnorm{\bW(t)}
    $ and $
 \abs{\Delta \bW_{ij}(t)} \leq \frac{\beta_2}{\sqrt{M}} \; %\max_{ij} \left \{\abs{\bW_{ij}(t)} \right\} 
 \maxnorm{\bV(t)}$
    % \end{align*}
    where 
$\beta_1 =\frac{4}{ \delta} c_lc_{in}Q\sqrt{Z} L_{\phi} $ and $\beta_2 = \frac{4}{ \delta} c_lc_{in}QDL_{\phi}$ are constants independent of $M$.
\begin{proof}
We first observe that the cosine similarity for a neural network $f(\cdot)$ as defined in \eqref{eqn: ntk_nn} can be written out as
\begin{align*}
    s(\bx_k , \bx_{k,q}) &= \frac{f(\bx_k )^\top f(\bx_{k,q})}{(\norm{f(\bx)}+\delta)(\norm{f(\by)} + \delta)} \\
    &= \frac{\phi^\top ( \bV\bx_k )\bW\bW^\top \phi( \bV\bx_{k,q})}{(\norm{\bW^\top \phi( \bV\bx_k )} + \sqrt{M} \delta) (\norm{\bW^\top \phi( \bV\bx_{k,q})} + \sqrt{M}\delta)}.
\end{align*}
To prove Lemma~\ref{lemma:change}, we split it into two parts. We first analyze $\frac{\partial \mathcal{L}}{\partial \bV}$ and then analyze $\frac{\partial \mathcal{L}}{\partial \bW}$.
\subsubsection*{Analysis for $\frac{\partial \mathcal{L}}{\partial \bV}$}
To bound $\frac{\partial \mathcal{L}}{\partial \bV}$, we first note that:
\begin{align*}
    \abs{\frac{\partial \mathcal{L}}{\partial \bV_{ij}}} &\leq \frac{1}{N}  \sum_{k=1}^{N} \sum_{q = 1}^{Q} \abs{\frac{\partial l}{\partial s(\bx_k ,\bx_{k,q})}} \abs{\frac{\partial s(\bx_k , \bx_{k,q})}{\partial \bV_{ij}}} 
\end{align*}
And note that $ \abs{\frac{\partial l}{\partial s(\bx_k ,\bx_{k,q})}}$ is bound by $c_l$ by Assumption~\ref{assm:smooth}. Therefore what remains is to bound the second term, $\abs{\frac{\partial s(\bx_k , \bx_{k,q})}{\partial \bV_{ij}}} $.
To do so, let us first note that $\bW^\top \phi(\bV\bx_k ) = \sqrt{M}f(\bx_k;\btheta )$ and $\bW^\top \phi(\bV\bx_{k,q}) = \sqrt{M}f(\bx_{k,q};\btheta)$. We denote $f(\bx_k ;\btheta)$ and $f(\bx_{k,q};\btheta)$ by $\bu_k $ and $\bu_{k,q} $ respectively.
Writing out $\frac{\partial l}{\partial s(\bx_k ,\bx_{k,q})}$, we obtain:
\begin{align}\label{eq:app:delSdelV}
    \frac{\partial s(\bx_k , \bx_{k,q})}{\partial \bV_{ij}} &= 
    \underbrace{\left(\frac{\partial s(\bx_k , \bx_{k,q})}{\partial \phi(\bV\bx_k )}\right)^\top \frac{\partial \phi(\bV\bx_k )}{\partial \bV_{ij}}}_{\text{Term I}} + 
    \underbrace{\left(\frac{\partial s(\bx_k , \bx_{k,q})}{\partial \phi(\bV\bx_{k,q})}\right)^\top \frac{\partial \phi(\bV\bx_{k,q})}{\partial \bV_{ij}}}_{\text{Term II}}
\end{align}
Observe that \emph{Term I} and \emph{Term II} only differ in considering $\bx_n$ and $\bx_{k,q}$. Therefore, we show the derivation for \emph{Term I} in the following - \emph{Term II} follow the same structure.

 \noindent To compute \emph{Term I}, we start with computing the term$\frac{\partial s(\bx_k , \bx_{k,q})}{\partial \phi(\bV\bx_k )}$ and obtain
\begin{align*}
    \frac{\partial s(\bx_k , \bx_{k,q})}{\partial \phi(\bV\bx_k )} &= \frac{1}{\sqrt{M}}\frac{(\norm{\bu_k } + \delta)(\norm{\bu_{k,q} } + \delta) \norm{\bu_k }\bW\bu_{k,q}  - \bu_k^\top \bu_{k,q}  (\norm{\bu_{k,q} } + \delta)\bW\bu_k }
    {(\norm{\bu_k }+\delta)^2(\norm{\bu_{k,q} } + \delta)^2\norm{\bu_k }}.
\end{align*}
% Similarly,
% \begin{align*}
%     \frac{\partial s(\bx_k , \bx_{k,q})}{\partial \phi(\bV\bx_{k,q})} &= \frac{1}{\sqrt{M}}\frac{(\norm{\bu_k } + \delta)(\norm{\bu_{k,q} } + \delta) \norm{\bu_{k,q} }\bW\bu_k  - \bu_k^\top \bu_{k,q}  (\norm{\bu_k } + \delta)\bW\bu_{k,q} }{(\norm{\bu_k }+\delta)^2(\norm{\bu_{k,q} } + \delta)^2\norm{\bu_{k,q} }}
% \end{align*}
It is easy to see that the vector $\frac{\partial \phi(\bV\bx_k )}{\partial \bV_{ij}}$ can be written as:
\begin{align*}
    \left[ \frac{\partial \phi(\bV\bx_k )}{\partial \bV_{ij}} \right]_r = \mathbf{1}_{r = i} \cdot \phi '(\bV_{i.}\bx_k )(\bx_k)_j
\end{align*}
and therefore we can write \emph{Term I} as
% where $\bV_{i.}$ denotes the $i^{\text{th}}$ row of $\bV$ and $x_{k}^j$ denotes the $j^{\text{th}}$ element of vector $\bx_k$ . Similarly,
% \begin{align*}
%     \left[ \frac{\partial \phi(\bV\bx_{k,q})}{\partial \bV_{ij}} \right]_r = \mathbf{1}_{r = i} \cdot \phi '(\bV_{i.}\bx_{k,q})(\bx_{kq})_j
% \end{align*}
\begin{align*}
   &{ \left(\frac{\partial s(\bx_k , \bx_{k,q})}{\partial \phi(\bV\bx_k )}\right)^\top \frac{\partial \phi(\bV\bx_k )}{\partial \bV_{ij}}} \\
   &=  \left[ \frac{\partial s(\bx_k , \bx_{k,q})}{\partial \phi(\bV\bx_{k,q})} \right]_i \cdot \phi '(\bV_{i.}\bx_k )(\bx_k)_j \\
    &= \frac{1}{\sqrt{M}}\frac{(\norm{\bu_k } + \delta)(\norm{\bu_{k,q} } + \delta) \norm{\bu_k }\bW_{i.}\bu_{k,q}  - {u_{k}^\top }\bu_{k,q}  (\norm{\bu_{k,q} } + \delta)\bW_{i.}\bu_k }{(\norm{\bu_k }+\delta)^2(\norm{\bu_{k,q} } + \delta)^2\norm{\bu_k }} \cdot \phi '(\bV_{i.}\bx_k )(\bx_k)_j.
\end{align*}
% And,
% \begin{align*}
%    &{ \left(\frac{\partial s(\bx_k , \bx_{k,q})}{\partial \phi(\bV\bx_{k,q})}\right)^\top \frac{\partial \phi(\bV\bx_{k,q})}{\partial \bV_{ij}}} \\
%    &=\frac{1}{\sqrt{M}}\frac{(\norm{\bu_k } + \delta)(\norm{\bu_{k,q} } + \delta) \norm{\bu_{k,q} }\bW_{i.}\bu_k  - {u_{k}^\top }\bu_{k,q}  (\norm{\bu_k } + \delta)\bW_{i.}\bu_{k,q} }{(\norm{\bu_k }+\delta)^2(\norm{\bu_{k,q} } + \delta)^2\norm{\bu_{k,q} }} \cdot \phi '(\bV_{i.}\bx_{k,q})(\bx_{kq})_j
% \end{align*}
From triangle inequality, we can bound the absolute value of \emph{Term I} as:
\begin{align*}
   &\abs{ \left(\frac{\partial s(\bx_k , \bx_{k,q})}{\partial \phi(\bV\bx_k )}\right)^\top \frac{\partial \phi(\bV\bx_k )}{\partial \bV_{ij}}} \\
   &\leq \frac{1}{\sqrt{M}}\frac{(\norm{\bu_k } + \delta)(\norm{\bu_{k,q} } + \delta) \norm{\bu_k }\abs{\bW^{i.}\bu_{k,q} } + \abs{{u_{k}^\top }\bu_{k,q} } (\norm{\bu_{k,q} } + \delta)\abs{\bW_{i.}\bu_k }}{(\norm{\bu_k }+\delta)^2(\norm{\bu_{k,q} } + \delta)^2\norm{\bu_k }} \\
   &\quad\cdot \abs{\phi '(\bV_{i.}\bx_k )(\bx_k)_j}.
\end{align*}
From Cauchy-Schwartz inequality, we know that $\bu_k ^\top \bu_{k,q}  \leq \norm{\bu_k } \norm{\bu_{k,q} }$, $\abs{\bW_{i.}\bu_k } \leq \norm{\bW_{i.}}\norm{\bu_k }$ and $\abs{\bW_{i.}\bu_{k,q} } \leq \norm{\bW_{i.}}\norm{\bu_{k,q} }$. The expression then can be simplified as:

\begin{align*}
  \abs{  \left(\frac{\partial s(\bx_k , \bx_{k,q})}{\partial \phi(\bV\bx_k )}\right)^\top \frac{\partial \phi(\bV\bx_k )}{\partial \bV_{ij}}} &\leq \frac{1}{\sqrt{M}} \frac{(2\norm{\bu_k } + \delta)\norm{\bu_{k,q} } \norm{\bW_{i.}}}{(\norm{\bu_k } + \delta)^2(\norm{\bu_{k,q} } + \delta)} \cdot \abs{\phi '(\bV_{i.}\bx_k )(\bx_k)_j} \\
    & \leq \frac{1}{\sqrt{M}} \frac{2\norm{\bW_{i.}}}{\norm{\bu_k } + \delta} \cdot \abs{\phi '(\bV_{i.}\bx_k )(\bx_k)_j} \\
    &\leq \frac{1}{\sqrt{M}} \frac{2\norm{\bW_{i.}}}{ \delta} \cdot \abs{\phi '(\bV_{i.}\bx_k )(\bx_k)_j}
\end{align*}
since $\delta > 0$. 

\noindent By the same argument we can bound \emph{Term II} as:
\begin{align*}
    \abs{ \left(\frac{\partial s(\bx_k , \bx_{k,q})}{\partial \phi(\bV\bx_{k,q})}\right)^\top \frac{\partial \phi(\bV\bx_{k,q})}{\partial \bV_{ij}} }&\leq \frac{1}{\sqrt{M}} \frac{2\norm{\bW_{i.}}}{ \delta} \cdot \abs{\phi '(\bV_{i.}\bx_{k,q})(\bx_k)_j}.
\end{align*}
Combining the bound for \emph{Term I} and \emph{Term II} in \eqref{eq:app:delSdelV}, we obtain:
\begin{align*}
     \abs{\frac{\partial s(\bx_k , \bx_{k,q})}{\partial \bV_{ij}}} &\leq \frac{1}{\sqrt{M}}  \frac{2 \norm{\bW_{i.}}}{\delta}  \left(  \abs{\phi '(\bV_{i.}\bx_k )(\bx_k)_j} + \abs{\phi '(\bV_{i.}\bx_{k,q})(\bx_k)_j} \right)   .
\end{align*}
Let $k'$ and $q'$ denote the values of $k$ and $q$ respectively at which $\abs{\frac{\partial l}{\partial s(\bx_k , \bx_{k,q})}} \abs{\frac{\partial s(\bx_k , \bx_{k,q})}{\partial \bV_{ij}}}$ is maximum. Then we have:
\begin{align*}
    \abs{\frac{\partial \mathcal{L}}{\partial \bV_{ij}}} &\leq Q \abs{\frac{\partial l}{\partial s(\bx_{k'} , \bx_{k',q'})}} \abs{\frac{\partial s(\bx_{k'} , \bx_{k',q'})}{\partial \bV_{ij}}} 
\end{align*}
Therefore:
\begin{align}
    \label{eqn:ep_w1}
    \abs{\frac{\partial \mathcal{L}}{\partial \bV_{ij}}} &\leq \frac{2Q}{\sqrt{M}}  \abs{\frac{\partial l}{\partial s(\bx_{k'} , \bx_{k',q'})}} \frac{ \norm{\bW_{i.}}}{ \delta}  \left(  \abs{\phi '(\bV_{i.}\bx_k )(\bx_{k})_j} + \abs{\phi '(\bV_{i.}\bx_{k,q})(\bx_{k,q})_j} \right)   
\end{align}
Under gradient descent with learning rate $\eta$, we have:
\begin{align*}
    \abs{\Delta \bV_{ij}} &= \eta  \abs{\frac{\partial \mathcal{L}}{\partial \bV_{ij}}} \\
    & \leq \frac{4Q\sqrt{Z}c_l L_{\phi} c_{in}}{\sqrt{M} \delta} \max_{ij} \left \{ \abs{\bW_{ij}} \right \}
\end{align*}
where we have used Assumption \ref{assm:smooth} to bound $\abs{\frac{\partial l}{\partial s(\bx_{k'} , \bx_{k',q'})}}\leq c_l$, \ref{assm:data} to bound $ \abs{(\bx_k)_j}, \abs{(\bx_{k,q})_j}\leq c_{in}$ and \ref{assm:activation} to bound $\abs{\phi '(\bV_{i.}\bx_k )}, \abs{\phi '(\bV_{i.}\bx_{k,q} )}$. Setting $\beta_1 =\frac{4}{ \delta} Q\sqrt{Z}c_l L_{\phi} c_{in}$, we have the required result.

\subsubsection*{Analysis for  $\frac{\partial \mathcal{L}}{\partial \bW}$}
The argument follows the one presented for $\frac{\partial \mathcal{L}}{\partial \bV}$ closely .
Note that:
\begin{align*}
    s(\bx_k , \bx_{k,q}) 
    &=  \frac{\text{Tr} 
 \left(\phi^\top (\bV\bx_k )\bW\bW^\top \phi(\bV\bx_{k,q})\right)}{(\norm{\bW^\top \phi(\bV\bx_k )} + \sqrt{M} \delta) (\norm{\bW^\top \phi(\bV\bx_{k,q})} + \sqrt{M}\delta)}
\end{align*}
Again, we can start by computing $ \frac{\partial s(\bx_k , \bx_{k,q})}{\partial \bW}$ as:
\begin{align*}
    \frac{\partial s(\bx_k , \bx_{k,q})}{\partial \bW} 
    &= \frac{1}{\sqrt{M}}\left[\frac{\phi(\bV\bx_k )\bu_{k,q} ^\top  + \phi(\bV\bx_{k,q})\bu_k ^\top }{(\norm{\bu_k } + \delta) (\norm{\bu_{k,q} } + \delta)} \right] \\
    &\quad- \frac{1}{\sqrt{M}}\left[ \frac{\bu_k ^\top \bu_{k,q} }{(\norm{\bu_k } + \delta)^2( \norm{\bu_{k,q} } + \delta)^2} \right.\\
   &\quad\quad\left. \left( \frac{(\norm{\bu_{k,q} } + \delta)\phi(\bV\bx_k ){\bu_k ^\top }}{ \norm{\bu_k }} + \frac{(\norm{\bu_k } + \delta)\phi(\bV\bx_{k,q}){\bu_{k,q} ^\top }}{ \norm{\bu_{k,q} }} \right) \right] 
\end{align*}
And therefore,
\begin{align*}
    \frac{\partial s(\bx_k , \bx_{k,q})}{\partial \bW_{ij}} &= \frac{1}{\sqrt{M}}\left[\frac{\phi^i(\bV\bx_k )\bu_{k,q} ^j + \phi^i(\bV\bx_{k,q})\bu_k ^j}{(\norm{\bu_k } + \delta) (\norm{\bu_{k,q} } + \delta)} \right] \\
    &\quad- \frac{1}{\sqrt{M}}\left[ \frac{\bu_k ^\top \bu_{k,q} }{(\norm{\bu_k } + \delta)^2( \norm{\bu_{k,q} } + \delta)^2} \right.\\
   &\quad\quad\left. \left( \frac{(\norm{\bu_{k,q} } + \delta)\phi_i(\bV\bx_k ){(\bu_k)_j}}{ \norm{\bu_k }} + \frac{(\norm{\bu_k } + \delta)\phi_i(\bV\bx_{k,q}){(\bu_{k,q})_j}}{ \norm{\bu_{k,q} }} \right) \right] 
\end{align*}
Using triangle and Cauchy-Schwartz inequalities, and the fact that $\delta > 0$, the expression can be simplified into:
\begin{align*}
\frac{\partial s(\bx_k , \bx_{k,q})}{\partial \bW_{ij}} &\leq \frac{1}{\sqrt{M}}\left[\frac{\abs{\phi_i(\bV\bx_k )} + \abs{\phi_i(\bV\bx_{k,q})}}{\delta} + \frac{1}{ \delta} \left( {\abs{\phi_i({\bV\bx_k })}} + {\abs{\phi_i(\bV\bx_{k,q})}} \right) \right] 
\end{align*}
Now,
\begin{align*}
\abs{\frac{\partial \mathcal{L}}{\partial \bW_{ij}}} &\leq \frac{1}{N}  \sum_{k=1}^{N} \sum_{q = 1}^{Q} \abs{\frac{\partial l}{\partial s(\bx_k , {\bx_{k,q})}}} \abs{\frac{\partial s(\bx_k , \bx_{k,q})}{\partial \bW_{ij}}}
    % \norm{\frac{\partial L}{\partial W_2}}_F &\leq \frac{1}{n}  \sum_{k=1}^{n} \sum_{q = 1}^{w} \abs{\frac{\partial l}{\partial s(\bx_k , \textbf{x$_\text{kq}$)}}} \norm{\frac{\partial s(\bx_k , \bx_{k,q})}{\partial W_2}}_F \\
\end{align*}
Let $k''$ and $q''$ denote the values of $k$ and $q$ respectively at which $\abs{\frac{\partial l}{\partial s(\bx_k , {\bx_{kq})}}} \abs{\frac{\partial s(\bx_k , \bx_{k,q})}{\partial \bW_{ij}}}$ is maximum. Then we have:
\begin{align*}
    \abs{\frac{\partial \mathcal{L}}{\partial \bW_{ij}}} &\leq \frac{2Q}{\sqrt{M}} \abs{\frac{\partial l}{\partial s({\bx_{k''}}, {\bx_{k''q''})}}} \left[ \frac{\abs{\phi_i(\bV\bx_k )} + \abs{\phi_i(\bV\bx_{k,q})}}{ \delta} \right] 
\end{align*}
Therefore, under gradient descent,
\begin{align*}
    \abs{\Delta \bW_{ij}} &= \eta  \abs{\frac{\partial \mathcal{L}}{\partial \bW_{ij}}} \\
    &\leq \frac{4QDc_lc_{in}L_{\phi}}{\sqrt{M} \delta} \max_{ij} \left \{ \abs{\bV_{ij}} \right \} 
\end{align*}
where we have again used Assumptions \ref{assm:smooth}, \ref{assm:data} and \ref{assm:activation}. Setting $\beta_2 = \frac{4}{ \delta} QDc_lc_{in}L_{\phi}$, we have the required result.
\end{proof}

\subsection{Lemma \ref{lemma:ct} [Bound on weight difference during training under cosine similarity]
}    
   Define $\beta := \max \{\beta_1, \beta_2\}$ and 
    $ c(t) := c(0)\left(1 + \frac{\beta}{\sqrt{M}} \right)^t $
    where $c(0) = c_{\theta} \log M$.
    % $ c(0) := \max\left\{\maxnorm{\bV(0)},\maxnorm{\bW(0)}\right\}$.
    Then, for any $t$, we have :
    % \begin{align*}
  $   \maxnorm{\bV(t) - \bV(0)} \leq  c(t) - c(0) $ and $
    \maxnorm{\bW(t) - \bW(0)} \leq  c(t) - c(0). $

\begin{proof} We prove the above statement by induction.

\paragraph{Base Case.} Note that at $t = 0$, these hold trivially:
\begin{align*}
 \max_{ij} \left \{\abs{\bV_{ij}(0) - \bV_{ij}(0)} \right \} &=0 \\
\max_{ij} \left \{\abs{\bW_{ij}(0) - \bW_{ij}(0)} \right \} &=0
\end{align*}

\paragraph{Induction Step.}
    Let us assume that at time $t$, we have:
\begin{align}
    \label{eqn:a2_w1} \max_{ij} \left \{\abs{\bV_{ij}(t) - \bV_{ij}(0)} \right \} &\leq  c(t) - c(0) \\
    \label{eqn:a2_w2} \max_{ij} \left \{\abs{\bW_{ij}(t) - \bW_{ij}(0)} \right \} &\leq  c(t) - c(0) 
\end{align}
Then, at $t+1$, we have:
\begin{align*}
    \max_{ij} \left \{\abs{\bV_{ij}(t+1) - \bV_{ij}(0)} \right \} &= \max_{ij} \left \{\abs{\bV_{ij}(t+1) - \bV_{ij}(t) + \bV_{ij}(t) - \bV_{ij}(0)} \right \} \\
    &= \max_{ij} \left \{\abs{\Delta \bV_{ij}(t) + \bV_{ij}(t) - \bV_{ij}(0)} \right \} \\
    &\leq  \max_{ij} \left \{\abs{\Delta \bV_{ij}(t)} + \abs{ \bV_{ij}(t) - \bV_{ij}(0)} \right \}
\end{align*}
where we have used the definition of $\Delta \bV_{ij}(t)$ (the change of $\bV_{ij}$ in one time-step at time $t$), triangle inequality and the fact that inequalities are preserved under $\max$ operation. Further, applying triangle inequality for the $\max$ operation, we have:
\begin{align}
    \max_{ij} \left \{\abs{\bV_{ij}(t+1) - \bV_{ij}(0)} \right \} &\leq \max_{ij} \left \{\abs{\Delta \bV_{ij}(t)} \right \} + \max_{ij} \left \{\abs{\bV_{ij}(t) - \bV_{ij}(0)} \right \} \nonumber\\
     \label{eqn:a2_w1_change} &\leq \frac{\beta_1}{\sqrt{M}} \max_{ij} \left \{\abs{\bW_{ij}(t)} \right\} + c(t) - c(0),
\end{align}
where we use  $   \abs{\Delta \bV_{ij}(t)} \leq \frac{\beta_1}{\sqrt{M}} \; %\max_{ij} \left \{\abs{\bV_{ij}(t)} \right\}
    \maxnorm{\bW(t)}
    $ from  Lemma~\ref{lemma:change} and \eqref{eqn:a2_w1} to obtain \eqref{eqn:a2_w1_change}.
Now by \eqref{eqn:a2_w2}, we have: 
\begin{align*}
    \max_{ij} \left \{\abs{\bW_{ij}(t)} - \abs{\bW_{ij}(0)} \right \} &\leq  c(t) - c(0) \\
    \implies \abs{ \max_{ij} \left \{\abs{\bW_{ij}(t)} \right \} -  \max_{ij} \left \{\abs{\bW_{ij}(0)} \right \}} &\leq  c(t) - c(0)
\end{align*}

\subsubsection*{Case 1: $\max_{ij} \left \{\abs{\bW_{ij}(t)} \right \} <  \max_{ij} \left \{\abs{\bW_{ij}(0)} \right \} $}

We start by multiplying $ \frac{\beta_1}{\sqrt{M}} $ to both sides of the inequality:
\begin{align*}
    \frac{\beta_1}{\sqrt{M}} \max_{ij} \left \{\abs{\bW_{ij}(t)} \right\} < \frac{\beta_1}{\sqrt{M}} \max_{ij} \left \{\abs{\bW_{ij}(0)} \right\}
\end{align*}
and then using (\ref{eqn:a2_w1_change}), we obtain:
\begin{align*}
    \max_{ij} \left \{\abs{\bV_{ij}(t+1) - \bV_{ij}(0)} \right \} &\leq   \frac{\beta_1}{\sqrt{M}} \max_{ij} \left \{\abs{\bW_{ij}(0)} \right\} + c(t) - c(0) 
\end{align*}
Note that $\max_{ij} \left \{\abs{\bW_{ij}(0)} \right\} \leq c(0)$ and $c(t) \geq c(0)$ by definition. Therefore, we have:
\begin{align*}
    \max_{ij} \left \{\abs{\bV_{ij}(t+1) - \bV_{ij}(0)} \right \} &\leq  \frac{\beta}{\sqrt{M}}c(t) + c(t) - c(0)\\
    &=  \left(1 + \frac{\beta}{\sqrt{M}} \right)c(t) - c(0)
\end{align*}
Now from the definition of $c(t)$, we know that $\left(1 + \frac{\beta}{\sqrt{m}} \right)c(t) = c(t+1)$ and therefore,
\begin{align*}
    \max_{ij} \left \{\abs{\bV_{ij}(t+1) - \bV_{ij}(0)} \right \} &\leq  c(t+1) - c(0)
\end{align*}

\subsubsection*{Case 2: $\max_{ij} \left \{\abs{\bW_{ij}(t)} \right \} \geq  \max_{ij} \left \{\abs{\bW_{ij}(0)} \right \} $}

We have:
\begin{align*}
    { \max_{ij} \left \{\abs{\bW_{ij}(t)} \right \} -  \max_{ij} \left \{\abs{\bW_{ij}(0)} \right \}} &\leq  c(t) - c(0) \\
    \implies \max_{ij} \left \{\abs{\bW_{ij}(t)} \right \}  &\leq  c(t) 
\end{align*}
since $\max_{ij} \left \{\abs{\bW_{ij}(0)} \right\} \leq c(0)$. Using this inequality in (\ref{eqn:a2_w1_change}), we have:
\begin{align*}
    \max_{ij} \left \{\abs{\bV_{ij}(t+1) - \bV_{ij}(0)} \right \} &\leq \frac{\beta_1}{\sqrt{M}} c(t) + c(t) - c(0) \\
    \implies \max_{ij} \left \{\abs{\bV_{ij}(t+1) - \bV_{ij}(0)} \right \} &\leq c(t+1) - c(0)
\end{align*}
By the same token, we have:
\begin{align*}
    \max_{ij} \left \{\abs{\bW_{ij}(t+1) - \bW_{ij}(0)} \right \} &\leq c(t+1) - c(0)
\end{align*}
Therefore, by the principle of induction, (\ref{eqn:a2_w1}) and (\ref{eqn:a2_w2}) hold for all $t$.
\end{proof}

\subsection{Theorem \ref{thm:ntk_conv}  [Bound on the change in NTK under cosine similarity]
}
    Consider losses of the form \eqref{eqn:loss} with \ref{eq:cos}. 
     Let $ c(0),\beta$ be the constant in Lemma~\ref{lemma:ct},   $R$ be as in \eqref{eq: R bound}\footnote{{Note that $R$ here is a function of $M$, with the relation being given by Lemma \ref{lemma:ct}. Similarly, $c(0) = c_{\theta} \log M$.}}, $\alpha_1, \alpha_2$ be as in Lemma \ref{lemma:hessian} and $\gamma:=2\sqrt{2}DL_{\phi}c_{in}$. If a neural network $f(\cdot)$ of the form \eqref{eqn: ntk_nn} is trained using gradient descent, then under Assumptions~\ref{assm:delta}~-~\ref{assm:gradient}, for $t \leq M^{\alpha}$ iterations, the change in NTK is bounded as 
     \begin{align*}
         \abs{\bK_{ij}(\bx , {\tilde\bx }; \btheta(t)) - \bK_{ij}(\bx , {\tilde\bx }; \btheta(0))} \leq \gamma  \left(c(0) e^{{\beta}{{M^{\alpha - 0.5}}}} \right) \frac{\alpha_1R^2 + \alpha_2R}{\sqrt{M}} .
        \end{align*}
    In particular, if we set $\alpha = \frac{1}{6}$ and assume $M \geq \max\{1, \beta^3\}$, then the above statement simplifies to
    \begin{align*}
        \max_{t \in \left(0, M^{1/6}\right]} \sup_{\bx, \Tilde{\bx}} \abs{\bK_{ij}(\bx , {\tilde\bx }; \btheta(t)) - \bK_{ij}(\bx , {\tilde\bx }; \btheta(0))} = O\left(M^{-1/6}(\log M)^3 \right).
    \end{align*}
\begin{proof}
The main result we need is Lemma~\ref{lemma:ntkhess}. Therefore, let us recall the statement of this lemma and highlight the quantities we need to specify to obtain Theorem~\ref{thm:ntk_conv}:

\noindent Define $\bbS := \{\bs \in \mathbb{R}^p; \norm{\bs - \bs(0)} \leq R\}$, where $p$ is the total number of learnable parameters in \eqref{eqn: ntk_nn}. Assume that for any input $\bx$, 
   \begin{enumerate}
       \item \label{item:hessian}$\norm{\bH^{(z)}(\bx; \bs)}_2 \leq \epsilon$ and 
        \item \label{item:gradient}$\norm{\nabla_{\bs}f_z({\bx; {\bs}})}_2 \leq c_0$, $\forall \; z \in [Z]$ and $\forall \; \bs \in \bbS$. 
        \end{enumerate}Then, for any inputs $\bx, \tilde\bx$, $\forall \; \bs \in \bbS$ and $\forall \; i,j \in [Z]$, $\abs{\bK_{ij}(\bx , {\tilde\bx }; {\bs}) - \bK_{ij}(\bx , {\tilde\bx }; {\bs(0)})} \leq 2\epsilon c_0 R$.
Let us now consider the two quantities  separately:
\subsubsection*{Hessian (Aspect \ref{item:hessian})}
Let us recall Lemma~\ref{lemma:hessian}:

\noindent   Under Assumptions~\ref{assm:data}, \ref{assm:activation}, \ref{assm:gradient}, consider the neural network defined in \eqref{eqn: ntk_nn}. If the change in weights during training is bounded as 
     \begin{align*}
         \norm{\bW(t) - \bW(0)}_F + \norm{\bV(t) - \bV(0)}_F \leq R,
         % \label{eq: R bound}
         \end{align*}
          then, $\forall \; z \in [Z]$, with $\alpha_1 = 4\beta_{\phi} c_{in}^2 L_{\phi}$ and $\alpha_2 = 4L_{\phi}c_{in}(1 + \beta_{\phi} c_{in}s_0 c_s)$,  the $z^\text{th}$ Hessian is bounded as:
   $  \norm{\bH^{(z)}(\bx ; \btheta(t))}_2 \leq \frac{\alpha_1 R + \alpha_2}{\sqrt{M}}.$

\noindent Now, let us assume we are training $f(\cdot)$ of the form \eqref{eqn: ntk_nn} for $M^{\alpha}$ iterations. Then, using Lemma \ref{lemma:ct}, we know that:
\begin{align*}
     \max_{ij} \left \{\abs{\bV_{ij}(M^{\alpha}) - \bV_{ij}(0)} \right \} &\leq  c(M^{\alpha}) - c(0)  \\
     \max_{ij} \left \{\abs{\bW_{ij}(M^{\alpha}) - \bW_{ij}(0)} \right \} &\leq  c(M^{\alpha}) - c(0) 
\end{align*}
and we can further bound the right hand side as:
\begin{align*}
    c(M^{\alpha}) - c(0) &=  c(0) \left[ \left(1 + \frac{\beta}{\sqrt{M}} \right)^{M^{\alpha}} -1 \right] \\
    &\leq c(0) \left[ \left(e^{{\beta}{{M^{\alpha - 0.5}}}} \right) -1 \right]
\end{align*}
where we have used the fact that $1 + x \leq e^x$. Therefore:
\begin{align*}
    \norm{\bV(M^{\alpha}) - \bV(0)}_F &\leq \sqrt{MD}  \max_{ij} \left \{\abs{\bV_{ij}(M^{\alpha}) - \bV_{ij}(0)} \right \} \\
    &\leq c(0)\sqrt{MD}  \left[ \left(e^{{\beta}{{M^{\alpha - 0.5}}}} \right) -1 \right]
\end{align*}
and similarly,
\begin{align*}
    \norm{\bW(M^{\alpha}) - \bW(0)}_F &\leq c(0)\sqrt{MZ}  \left[ \left(e^{{\beta}{{M^{\alpha - 0.5}}}} \right) -1 \right]
\end{align*}
We can now invoke Lemma~\ref{lemma:hessian} with $R = c(0)\sqrt{M}(\sqrt{D} + \sqrt{Z})  \left[ \left(e^{{\beta}{{M^{\alpha - 0.5}}}} \right) -1 \right]$.

\subsubsection*{Gradient (Aspect \ref{item:gradient})}
To further use Lemma~\ref{lemma:ntkhess}, we would need bounds on $\norm{\nabla_\btheta f_z(\bx;\btheta)}$ and  $ \norm{\nabla_\btheta f_z(\by;\btheta)}$, where $\btheta$ is the weight vector at $t = M^{\alpha}$. Now,
\begin{align*}
    \norm{\nabla_\btheta f_z(\bx;\btheta)}_2 &= \sqrt{\norm{\frac{\partial f_z(\bx;\btheta)}{\partial \bV}}^2_F + \norm{\frac{\partial f_z(\bx;\btheta)}{\partial \bW}}^2_F}
\end{align*}
Note that:
\begin{align*}
    \frac{\partial f_z(\bx;\btheta)}{\partial {\bV_{ij}}} &= \frac{1}{\sqrt{M}} \bW_{iz}\phi'({\bV_{i.}} \bx ) \bx_j \\
    \implies \abs{\frac{\partial f_z(\bx;\btheta)}{\partial {\bV_{ij}}}} &\leq \frac{1}{\sqrt{M}} \abs{\bW_{iz}} L_{\phi} \abs{x_j} \\
    &\leq \frac{ L_{\phi}  c_{in}}{\sqrt{M}} \left(c(0) \left[ \left(e^{{\beta}{{M^{\alpha - 0.5}}}} \right) -1 \right]+ \abs{\bW_{zi}(0)} \right) \\
    &\leq \frac{ L_{\phi}   c_{in} c(0)}{\sqrt{M}}   \left(e^{{\beta}{{M^{\alpha - 0.5}}}} \right)
\end{align*}
where we have used Lemma \ref{lemma:ct} and the fact that $\max_{ij} \left \{ \abs{\bW_{ij}(0)} \right  \} \leq c(0)$. Therefore,
\begin{align*}
    \norm{\frac{\partial f_z(\bx;\btheta)}{\partial \bV}}_F \leq  { L_{\phi} \sqrt{D}  c_{in} c(0)}  \left(e^{{\beta}{{M^{\alpha - 0.5}}}} \right)
\end{align*}
Similarly,
\begin{align*}
    \frac{\partial f_z(\bx;\btheta)}{\partial {\bW_{ij}}} &= \frac{1}{\sqrt{M}} \phi(\bV_{i.}\bx) \cdot \mathbf{1}_{j = z} \\
    &\leq \frac{L_{\phi}}{\sqrt{M}} \norm{{\bV_{i.}}}\norm{\bx} \cdot \mathbf{1}_{j = z} \\
    &\leq \frac{L_{\phi} {D} c_{in}}{\sqrt{M}} \max_{i j}\abs{V_{ij}} \cdot \mathbf{1}_{j = z} \\
    \implies \norm{\frac{\partial f_z(\bx;\btheta)}{\partial \bW}}_F &\leq L_{\phi} {D} c_{in} c(0) \left(e^{{\beta}{{M^{\alpha - 0.5}}}} \right) 
\end{align*}
Therefore,
\begin{align*}
     \norm{\nabla_\btheta f_z(\bx;\btheta)}_2 &\leq \sqrt{2}DL_{\phi} c_{in} c(0) \left(e^{{\beta}{{M^{\alpha - 0.5}}}} \right)
\end{align*}
Finally, we can use the bound on the Hessian and the gradient in Lemma~\ref{lemma:ntkhess} to bound the change in the tangent kernel:
\begin{align*}
\abs{K_{ij}(\bx, \by; \bs') - K_{ij}(\bx, \by; \bs(0)  )} &\leq \gamma  \left(c(0) e^{{\beta}{{M^{\alpha - 0.5}}}} \right) \frac{\alpha_1R^2 + \alpha_2R}{\sqrt{M}} 
\end{align*}
where $\gamma = 2\sqrt{2}DL_{\phi}c_{in}$ is a positive constant independent of $M$. 

\paragraph{Proof of second statement.} Now, let us consider the specific case where $\alpha = \frac{1}{6}$ and $M \geq \max\{1, \beta^3\}$ and look at the dependence of the change in Kernel on $M$. Note that:
\begin{align}
    \label{eq:app:o_gen}
    \abs{K_{ij}(\bx, \by; \bs') - K_{ij}(\bx, \by; \bs(0)  )} &= O\left( c(0) e^{{\beta}{{M^{{-1}/{3}}}}} \left(\frac{\alpha_1R^2 + \alpha_2R}{\sqrt{M}}\right)  \right)
\end{align}
since $\gamma$ is independent of $M$. Now, we have:
\begin{align}
    \label{eq:app:o_c}
    c(0) = O(\log M)
\end{align}
Further, note that 
\begin{align*}
    \beta M^{{-1}/{3}} \leq 1
\end{align*}
since $M \geq \max\{1, \beta^3\}$. Note the fact that when $z \leq 1$,
\begin{align}
    e^z &\leq 1 + 2z \leq 3 \nonumber\\
   \label{eq:app:exp_sm} \implies e^{\beta M^{{-1}/{3}}} &\leq 1 + 2\beta M^{{-1}/{3}} \leq 3
\end{align}
Using \eqref{eq:app:exp_sm}, we have:
\begin{align}
    \label{eq:app:o_exp}e^{{\beta}{{M^{{-1}/{3}}}}} &= O(1)
\end{align}
Now:
\begin{align*}
    \alpha_2 R = \alpha_2 (\sqrt{D} + \sqrt{Z}) c(0)\sqrt{M}  \left[ \left(e^{{\beta}{{M^{\alpha - 0.5}}}} \right) -1 \right]
\end{align*}
Using the fact that $\alpha_2, D, Z$ are independent of M, we have:
\begin{align}
    \label{eq:app:o_r}
    \alpha_2 R &= O(\sqrt{M}\log M \left[ \left(e^{{\beta}{{M^{\alpha - 0.5}}}} \right) -1 \right]) \nonumber\\
    &= O((\sqrt{M}\log M) M^{{-1}/{3}}) \nonumber\\
    &= O(M^\frac{1}{6} \log M )
\end{align}
where we have used \eqref{eq:app:o_c} and \eqref{eq:app:exp_sm}. Hence,
\begin{align}
    \label{eq:app:o_r2}
    \alpha_1 R^2 &=  O(M^\frac{1}{3} (\log M)^2 )
\end{align}
Plugging \eqref{eq:app:o_c}, \eqref{eq:app:o_exp}, \eqref{eq:app:o_r} and \eqref{eq:app:o_r2} into \eqref{eq:app:o_gen}, we have:
\begin{align*}
    \abs{K_{ij}(\bx, \by; \bs') - K_{ij}(\bx, \by; \bs(0)  )} &= O\left(\frac{(\log M)^3}{M^{\frac{1}{6}}} \right)
\end{align*}

\end{proof}

\subsection{Lemma \ref{lemm:cw_change} [Constancy of $C_V(t)$]}
  Under  Assumptions~\ref{assm:delta}--\ref{assm:activation} and constraint $\bW^{\top}\bW = \bbI_Z$, consider training $f^{\perp}(\cdot)$ in \eqref{eq:orth_nn} for $t$ iterations using Grassmannian gradient descent\footnote{In short, following \cite{edelman1998geometry}, the derivative of a function $g(\cdot)$ restricted to a Grassmannian manifold can be obtained by left-multiplying $1-g(\cdot) g(\cdot)^\top$ to the unrestricted derivative of $g(\cdot)$.}under losses of the form \eqref{eq:C_v} with learning rate $\eta$. Then 
    %\begin{align*}
    $ \norm{ \bC_V(t) - \bC_V(0) }_{F} \leq \kappa \frac{t}{\sqrt{M}}$,
    where $\kappa := 16{\delta^{-2}}\eta Q^2L_{\phi}^2c_{in}^2\sqrt{D}$.
    %    \end{align*}

    % Under  Assumptions~\ref{assm:delta}~-~\ref{assm:activation} and the constraint $\bW^{\top}\bW = \bbI_Z$, consider training $f^{\perp}(\cdot)$ of the form \eqref{eq:orth_nn} for $t$ iterations using Grassmannian gradient descent under losses of the form \eqref{eq:C_v} with learning rate $\eta$. Then, if $\kappa := 16{\delta^{-2}}\eta Q^2L_{\phi}^2c_{in}^2\sqrt{D}$, we have
    % %\begin{align*}
    % $ \norm{ \bC_V(t) - \bC_V(0) }_{F} \leq \kappa \frac{t}{\sqrt{M}}$.
    % %    \end{align*}
\begin{proof}
To bound the change in ${\bC_{\bV}}$ after $t$ iterations, we analyze $\norm{\bC_V(t) - \bC_V(0)}_F$. Recall that:
\begin{align}
    \label{eqn:def_C}
    \bC_V(t) &= \frac{1}{N} \sum_{n = 1}^N \sum^Q_{q = 1}\left( {\alpha_q} s(\bx_n,\bx_{nq}; t)\right)
\end{align}
where
\begin{align*}
    s(\bx, \by; t) &:=  \frac{ \phi(\bV(t)\bx)\phi^\top (\bV(t)\by)}{(\norm{\phi(\bV(t)\bx)} + \delta')(\norm{\phi(\bV(t)\by)} + \delta')}
\end{align*}
% Similarly,
% \begin{align*}
%     \bC_V(0) &= \frac{1}{N} \sum_{n = 1}^N \sum^Q_{q = 1}\left( {\alpha_q} s(\bx_n,\bx_{nq}; 0)\right)
% \end{align*}
and  $\delta':=\sqrt{M}\delta$. Therefore:
\begin{align}
    \label{eqn:cv_change}
    \norm{\bC_V(t) - \bC_V(0)}_F &\leq \frac{1}{N}\sum_{n = 1}^N \sum^Q_{q = 1} \left[ \norm{ s(\bx_n,\bx_{n,q}; t) - s(\bx_n, \bx_{n,q}; 0)}_F  \right]
\end{align}
where we have used the triangle inequality and the fact that $\abs{\alpha_q} = 1$. Now,
\begin{align*}
    &{ s(\bx, \by; t) - s(\bx, \by; 0)} \\
    &=  \frac{ \phi(\bV(t)\bx)\phi^\top (\bV(t)\by)}{(\norm{\phi(\bV(t)\bx)} + \delta')(\norm{\phi(\bV(t)\by)} + \delta')} 
    - \frac{ \phi(\bV(0)\bx)\phi^\top (\bV(0)\by)}{(\norm{\phi(\bV(0)\bx)} + \delta')(\norm{\phi(\bV(0)\by)} + \delta')}  \\
    &= \frac{\phi(\bV(t)\bx)}{\norm{\phi(\bV(t)\bx)} + \delta'}\left(\frac{\phi(\bV(t)\by)}{\norm{\phi(\bV(t)\by)} + \delta'} - \frac{\phi(\bV(0)\by)}{\norm{\phi(\bV(0)\by)} + \delta'}  \right)^\top  \\
    &\hspace*{1cm} +  \left(\frac{\phi(\bV(t)\bx)}{\norm{\phi(\bV(t)\bx)} + \delta'} - \frac{\phi(\bV(0)\bx)}{\norm{\phi(\bV(0)\bx)} + \delta'}  \right)\frac{\phi^\top (\bV(0)\by)}{\norm{\phi(\bV(0)\by)} + \delta'}
\end{align*}
On computing the Frobenius  norm of the above expression, we have:
\begin{align*}
   & \norm{s(\bx, \by; t) - s(\bx, \by; 0)}_F \\
   &\leq \frac{\norm{\phi(\bV(t)\bx)}}{\norm{\phi(\bV(t)\bx)} + \delta'}\norm{\frac{\phi(\bV(t)\by)}{\norm{\phi(\bV(t)\by)} + \delta'} - \frac{\phi(\bV(0)\by)}{\norm{\phi(\bV(0)\by)} + \delta'}} \\
    &\hspace*{1cm} +  \norm{\frac{\phi(\bV(t)\bx)}{\norm{\phi(\bV(t)\bx)} + \delta'} - \frac{\phi(\bV(0)\bx)}{\norm{\phi(\bV(0)\bx)} + \delta'}}\frac{\norm{\phi(\bV(0)\by)}}{\norm{\phi(\bV(0)\by)} + \delta'} \\
    &\leq \norm{\frac{\phi(\bV(t)\by)}{\norm{\phi(\bV(t)\by)} + \delta'} - \frac{\phi(\bV(0)\by)}{\norm{\phi(\bV(0)\by)} + \delta'}}   +  \norm{\frac{\phi(\bV(t)\bx)}{\norm{\phi(\bV(t)\bx)} + \delta'} - \frac{\phi(\bV(0)\bx)}{\norm{\phi(\bV(0)\bx)} + \delta'}}
\end{align*}
Now, we have:
\begin{align*}
    &\norm{\frac{\phi(\bV(t)\by)}{\norm{\phi(\bV(t)\by)} + \delta'} - \frac{\phi(\bV(0)\by)}{\norm{\phi(\bV(0)\by)} + \delta'}}\\
    &\leq \norm{\frac{\norm{\phi(\bV(0)\by}\phi(\bV(t)\by) - \norm{\phi(\bV(t)\by)}\phi(\bV(0)\by) }{(\norm{\phi(\bV(t)\by)} + \delta')(\norm{\phi(\bV(0)\by)} + \delta')}} \\
    &\hspace*{0.3cm} + \delta' \norm{\frac{\norm{\phi(\bV(t)\by) - \phi(\bV(0)\by)}}{(\norm{\phi(\bV(t)\by)} + \delta')(\norm{\phi(\bV(0)\by)} + \delta')}}
\end{align*}
Now,
\begin{align*}
    &\norm{ \left(\norm{\phi(\bV(0)\by}\phi(\bV(t)\by) - \norm{\phi(\bV(t)\by)}\phi(\bV(0)\by) \right)} 
    \\&\leq \abs{ \left(\norm{\phi(\bV(0)\by)} - \norm{\phi(\bV(t)\by)} \right)}\norm{\phi(\bV(t)\by)}  + \norm{\phi(\bV(t)\by}\norm{\phi(\bV(t)\by) - \phi(\bV(0)\by)}
\end{align*}
From the Lipschitz continuity of $\phi$, we have:
\begin{align*}
   \abs{ \left(\norm{\phi(\bV(t)\by)} - \norm{\phi(\bV(0)\by)} \right)} \leq \norm{\phi(\bV(t)\by) - \phi(\bV(0)\by)} &\leq L_{\phi}\norm{\bV(t) - \bV(0)}\norm{\by}
\end{align*}
Therefore,
\begin{align*}
    \norm{ \left(\norm{\phi(\bV(0)\by}\phi(\bV(t)\by) - \norm{\phi(\bV(t)\by)}\phi(\bV(0)\by) \right)} &\leq 2L_{\phi}\norm{\phi(\bV(t)\by)}\norm{\bV(t) - \bV(0)}\norm{\by}
\end{align*}
Hence,
\begin{align*}
    \norm{\frac{\phi(\bV(t)\by)}{\norm{\phi(\bV(t)\by)} + \delta'} - \frac{\phi(\bV(0)\by)}{\norm{\phi(\bV(0)\by)} + \delta'}}&\leq \frac{L_{\phi}\norm{\bV(t) - \bV(0)}\norm{\by}(2\norm{\phi(\bV(t)\by)} + \delta')}{(\norm{\phi(\bV(t)\by)} + \delta')(\norm{\phi(\bV(0)\by)} + \delta')} \\
    &\leq \frac{2L_{\phi}\norm{\bV(t) - \bV(0)}\norm{\by}}{\norm{\phi(\bV(0)\by)} + \delta'}
\end{align*}
Similarly,
\begin{align*}
    \norm{\frac{\phi(\bV(t)\bx)}{\norm{\phi(\bV(t)\bx)} + \delta'} - \frac{\phi(\bV(0)\bx)}{\norm{\phi(\bV(0)\bx)} + \delta'}} &\leq \frac{2L_{\phi}\norm{\bV(t) - \bV(0)}\norm{\bx}}{\norm{\phi(\bV(0)\bx)} + \delta'}
\end{align*}
Therefore,
\begin{align}
    \label{eqn:s_change}
    &\norm{s(\bx, \by; t) - s(\bx, \by; 0)}_F \nonumber\\
    &\leq 2L_{\phi}\norm{\bV(t) - \bV(0)} \left( \frac{\norm{\by}}{\norm{\phi(\bV(0)\by)} + \delta'} + \frac{\norm{\bx}}{\norm{\phi(\bV(0)\bx)} + \delta'} \right)
\end{align}

\subsubsection*{Bounding $\norm{\bV(t) - \bV(0)} $:}
We will derive $\norm{\bV(t) - \bV(0)} $ through the analysis of $\frac{\partial\calL}{\partial\bV}$. Recall that we can write the loss as :
\begin{align*}
    \mathcal{L} &= \frac{1}{N} \sum_{n = 1}^N \sum_{q=1}^Q\text{Tr}(\bW^T s({\bx_{n}},\bx_{n,q}) \bW ) 
\end{align*}
Let us define $g(\bx, \by) = \text{Tr}(\bW^T s(\bx, \by) \bW )$ and note that,
\begin{align*}
    \frac{\partial \mathcal{L}}{\partial \bV_{ij}} &= \frac{1}{N} \sum_{n = 1}^N   \sum_{q=1}^Q \frac{\partial g({\bx_{n}},\bx_{n,q})}{ \partial \bV_{ij}} 
\end{align*}
% We have:
% \begin{align*}
%     \mathcal{L} &= \frac{-1}{N} \sum_{n = 1}^N \left[ \text{Tr}(\textbf{W$_2^\top $} s(\textbf{x$_n^+$}, \textbf{x$_n$}) \bW ) + \text{Tr}(\textbf{W$_2^\top $} 
%  s(\textbf{x$_n^-$}, \textbf{x$_n$}) \bW) \right]
% \end{align*}
% Let us define $g(\bx, \by) = \text{Tr}(\textbf{W$_2^\top $} s(\bx, \by) \bW )$.  Note that,
% \begin{align*}
%     \frac{\partial \mathcal{L}}{\partial \bV_{ij}} &= \frac{-1}{N} \sum_{n = 1}^N \left[  \frac{\partial g(\textbf{x$_n^+$}, \textbf{x$_n$})}{ \partial \bV_{ij}} + \frac{\partial g(\textbf{x$_n^-$}, \textbf{x$_n$})}{\ \partial \bV_{ij}} \right]
% \end{align*}
Now,
\begin{align*}
    \frac{\partial g(\bx, \by)}{\partial \bV_{ij}} &= 
   \underbrace{ \left(\frac{\partial g(\bx, \by)}{\partial \phi(\bV\bx)}\right)^\top \frac{\partial \phi(\bV\bx)}{\partial \bV_{ij}}}_{\text{Term I}} + 
   \underbrace{ \left(\frac{\partial g(\bx, \by)}{\partial \phi(\bV\by)}\right)^\top \frac{\partial \phi(\bV\by)}{\partial \bV_{ij}}}_{\text{Term II}}
\end{align*}
We bound \emph{Term I} (\emph{Term II} follows analogously).
First, we expand the first expression in \emph{Term I}:
\begin{align*}
    \frac{\partial g(\bx, \by)}{\partial \phi(\bV\bx)} &=  \frac{\sqrt{M}\left(\norm{\phi(\bV\bx)} + \delta'\right)^2 \bW \bu_y - M^2 \bu_x^{\top}\bu_y\phi(\bV\bx) }{\left(\norm{\phi(\bV\bx)} + \delta'\right)^3\left(\norm{\phi(\bV\by)} + \delta'\right)}
\end{align*}
where ${\bu_x} = \frac{1}{\sqrt{M}} {\bW^\top }\phi({\bV\bx})$ and $\bu_y = \frac{1}{\sqrt{M}} {\bW^\top }\phi({\bV\by})$ and the second expression in \emph{Term I} as
\begin{align*}
    \left[ \frac{\partial \phi(\bV\bx)}{\partial \bV_{ij}} \right]_r = \mathbf{1}_{r = i} \cdot \phi '(\bV_{i.}\bx)\bx_j.
\end{align*}
Combining them, \emph{Term I} becomes:
\begin{align*}
    \left(\frac{\partial g(\bx, \by)}{\partial \phi(\bV\bx)}\right)^\top \frac{\partial \phi(\bV\bx)}{\partial \bV_{ij}} &= \frac{\sqrt{M}\left(\norm{\phi(\bV\bx)} + \delta'\right)^2 {\bW_{i.}} {\bu_y} - M{\bu_x^\top }{\bu_y}\phi_i(\bV\bx) }{\left(\norm{\phi(\bV\bx)} + \delta'\right)^3\left(\norm{\phi(\bV\by)} + \delta'\right)} \cdot \phi '(\bV_{i.}\bx)\bx_j
\end{align*}
We can bound its norm as:
\begin{align*}
   & \abs{ \left(\frac{\partial g(\bx, \by)}{\partial \phi(\bV\bx)}\right)^\top \frac{\partial \phi(\bV\bx)}{\partial \bV_{ij}}} \\
    &\leq \frac{\sqrt{M}\left(\norm{\phi(\bV\bx)} + \delta'\right)^2 \norm{{\bW_{i.}}}\norm{{\bu_y}} + M\norm{{\bu_x}}\norm{{\bu_y}}\abs{\phi_i(\bV\bx)} }{\left(\norm{\phi(\bV\bx)} + \delta'\right)^3\left(\norm{\phi(\bV\by)} + \delta'\right)} \cdot \abs{\phi '(\bV_{i.}\bx)\bx_j}.
\end{align*}
To further simplify the expression, we note that:
\begin{align*}
    \sqrt{M}\norm{{\bu_x}} &\leq \norm{\bW}_2\norm{\phi({\bV\bx})} \\
    &\leq (\norm{\phi(\bV\bx)} + \delta') \\
    \sqrt{M}\norm{{\bu_y}} &\leq  (\norm{\phi(\bV\by)} + \delta') \\
    \abs{\phi^i(\bV\bx)} &\leq \norm{\phi(\bV\bx)} + \delta'\\
        \norm{{\bW_{i.}}} &\leq 1 
\end{align*}
where the last inequality follows from the fact that $\bW^{\top}\bW = \bbI_Z$ (To see this, consider the $M \times M$ square matrix $\Tilde{\bW}$ formed by appending $M - Z$ columns to $\bW$ such that columns of $\Tilde{\bW}$ form an orthonormal set. Then, it is clear that $\Tilde{\bW}$ is an orthogonal square matrix and hence $\norm{{\Tilde{\bW}_{i.}}} = 1$. Since $\norm{{\bW_{i.}}} \leq \norm{{\Tilde{\bW}_{i.}}}$ by construction, we have $\norm{{\bW_{i.}}} \leq 1$). Therefore,
\begin{align*}
   \abs{ \left(\frac{\partial g(\bx, \by)}{\partial \phi(\bV\bx)}\right)^\top \frac{\partial \phi(\bV\bx)}{\partial \bV_{ij}} } &\leq  \frac{2\abs{\phi '(\bV_{i.}\bx)\bx_j}}{{\norm{\phi(\bV\bx)} + \delta'}} \\
   &\leq \frac{2}{\sqrt{M}}\left( \frac{L_{\phi}c_{in}}{\delta} \right)
\end{align*}
Similarly, we can bound \emph{Term II} as:
\begin{align*}
    \abs{ \left(\frac{\partial g(\bx, \by)}{\partial \phi(\bV\by)}\right)^\top \frac{\partial \phi(\bV\by)}{\partial \bV_{ij}} } &\leq \frac{2}{\sqrt{M}}\left( \frac{L_{\phi}c_{in}}{\delta} \right).
\end{align*}
Combining the bounds for \emph{Term I} and \emph{Term II}, we can now bound $\abs{\frac{\partial g(\bx, \by)}{\partial \bV_{ij}}}$ as:
\begin{align*}
     \abs{\frac{\partial g(\bx, \by)}{\partial \bV_{ij}}} &\leq \frac{4}{\sqrt{M}}\left( \frac{L_{\phi}c_{in}}{\delta} \right)
\end{align*}
and hence,
\begin{align*}
    \abs{\frac{\partial \mathcal{L}}{\partial \bV_{ij}}} &\leq \frac{4}{\sqrt{M}}\left( \frac{QL_{\phi}c_{in}}{\delta} \right).
\end{align*}
Assuming gradient descent with learning rate $\eta$ we have:
\begin{align*}
    \abs{\Delta \bV_{ij}} &= \eta  \abs{\frac{\partial \mathcal{L}}{\partial \bV_{ij}}} \\
 &\leq \frac{4}{\sqrt{M}}\left( \frac{\eta Q L_{\phi}c_{in}}{\delta} \right)
\end{align*}
Therefore, after $t$ iterations, the change in $\bV_{ij}$ can be bounded as:
\begin{align}
    \label{eqn:ch_v}
    \abs{\bV_{ij}(t) - \bV_{ij}(0)} &\leq \frac{t}{\sqrt{M}}\left( \frac{4\eta QL_{\phi}c_{in}}{\delta} \right) \nonumber\\
    \implies \norm{\bV(t) - \bV(0)}_F &\leq t\left( \frac{4\eta QL_{\phi}c_{in}}{\delta} \right).
\end{align}
Substituting this in (\ref{eqn:s_change}), we obtain:
\begin{align*}
    \norm{s(\bx, \by; t) - s(\bx, \by; 0)}_F &\leq t \left( \frac{4\eta QL_{\phi}c_{in}}{\delta} \right)\left( \frac{4 L_{\phi} {c_{in}} \sqrt{D}}{\sqrt{M} \delta} \right),
\end{align*}
and hence from (\ref{eqn:cv_change}), with $\kappa = \frac{16\eta Q^2L_{\phi}^2c_{in}^2\sqrt{D}}{\delta^2}$, we have:
\begin{align*}
    \norm{\bC_V(t) - \bC_V(0)}_F &= \kappa \frac{t}{\sqrt{M}} 
\end{align*}
therefore concluding the proof.
\end{proof}

\subsection{Lemma \ref{lemma:rep_close} [Perturbation bound on representation]}
Let $u(\textbf{x};\bW^*,\bC_\vartheta)$ be the representation obtained from \eqref{eq:orth_nn} with $\bW = \bW^*$, where $\bW^*$ is obtained by solving \eqref{eq: PCA}  for a fixed $\bC_\vartheta$. Under Assumptions~\ref{assm:delta}--\ref{assm:activation}, let $\bW^*,\widetilde\bW^*$ be the solutions of \eqref{eq: PCA} obtained through PCA on fixed $\widetilde\bC_V(0)$ and $\widetilde\bC_V(t)$ respectively. Let $\lambda_{Z},\lambda_{Z+1}$ be $Z^{\text{th}}$ and $(Z+1)^{\text{th}}$ eigenvalues of $\widetilde\bC_V(0)$. Let $\zeta = 4\delta^{-1}\eta Q\sqrt{D} L_{\phi}^2c_{in}^2$ and $\xi = 2^{\frac{7}{2}}\delta^{-1} QDc_{in}(L_{\phi}c_{\theta} + \abs{\phi(0)})$. There exists an orthogonal matrix $\bO$ such that 
\begin{align*}
\norm{\bO^\top u(\textbf{x};\widetilde\bW^*,\widetilde\bC_V(t)) - {u}(\textbf{x};\bW^*,\widetilde\bC_V(0))} \leq \zeta\frac{t}{\sqrt{M}}\left(1 + \frac{\xi \log M}{\lambda_{Z} - \lambda_{Z+1}} \right).
\end{align*}

\begin{proof}
    Note that since we are doing PCA, $\bW^\top \bW= \bbI_Z$ and hence the requirements of Lemma \ref{lemm:cw_change} are met. Therefore:
\begin{align}
\label{eqn:lem}
    \norm{\bC_V(t)- \bC_V(0)}_F &= \kappa \frac{t}{\sqrt{M}} \nonumber\\
    \implies \norm{\widetilde\bC_V(t)- \widetilde\bC_V(0)}_F &= \kappa \frac{t}{\sqrt{M}}
\end{align}
From Assumptions~\ref{assm:w_init} and \ref{assm:activation}, we have:
\begin{align}
    \label{eqn:ph1}
    \norm{\phi(\bV(0))}_F &\leq L_{\phi} \norm{\bV(0)}_F + \norm{\phi(\mathbf{0})}_F\nonumber \\
    \implies \norm{\phi(\bV(0))}_F &\leq (\sqrt{D}L_{\phi} c_{\theta})(\sqrt{M}\log M ) + \abs{\phi(0)}\sqrt{DM} \\
    &\leq \sqrt{D}(L_{\phi} c_{\theta} + \abs{\phi(0)})(\sqrt{M}\log M )
\end{align}
Using (\ref{eqn:ch_v}), we have:
\begin{align}
    \label{eqn:ph2}
    \norm{\phi(\bV(t)) - \phi(\bV(0))}_F &\leq L_{\phi} \norm{\bV(t) - \bV(0)}_F \nonumber\\
    &\leq \left( \frac{4\eta QL_{\phi}^2c_{in}}{\delta} \right)t
\end{align}
% Furthermore, from the definition of $\bC_V(0) = \frac{1}{N} \sum_{n = 1}^N \sum^Q_{q = 1}\left( {\alpha_q} s(\bx_n,\bx_{nq}; 0)\right)$ in (\ref{eqn:def_C}) using the fact $s(\bx, \by; 0) \leq 1$:
% \begin{align}
%     \label{eqn:ph3}
%     \norm{\bC_V(0)}_F \leq Q.
% \end{align}
 We denote $u(\bx;\widetilde\bW^*,\widetilde\bC_V(t))$ as $u$ and $\Tilde{u}(\bx;\bW^*,\widetilde\bC_V(0))$ as $\Tilde{u}$ to simplify notation. Now, for any orthogonal matrix $\bO$, using (\ref{eqn:ph1}) and (\ref{eqn:ph2}), we have:
\begin{align*}
    \norm{\bO^\top\bu - \Tilde{\bu}} &= \norm{\frac{1}{\sqrt{M}}\left( \bO^\top\widetilde\bW^{*^\top}\phi(\bV(t)\bx) - \bW^{*^\top}\phi(\bV(0)\bx) \right)} \\
    &\leq \frac{1}{\sqrt{M}} \left( \norm{ \bO^\top\widetilde\bW^{*^\top} \left( \phi(\bV(t)\bx) - \phi(\bV(0)\bx) \right)} + \norm{\left( \widetilde\bW^*\bO - \bW^*\right)^\top \phi(\bV(0)\bx)  }\right) \\
    &\leq \frac{1}{\sqrt{M}} \left( \norm{\bx}\norm{\phi(\bV(t)) - \phi(\bV(0))}_F +  \norm{\bx}\norm{\phi(\bV(0))}_F\norm{\widetilde\bW^*\bO - \bW^*}_F  \right)\\
    &\leq \frac{c_1t}{\sqrt{M}} + (c_2 \log M) \norm{\widetilde\bW^*\bO - \bW^*}_F 
\end{align*}
where $c_1 = \frac{4\eta Q\sqrt{D} L_{\phi}^2c_{in}^2}{\delta}$ and $c_2 = Dc_{in}(L_{\phi}c_{\theta} + \abs{\phi(0)})$. Now, note that $\widetilde\bW^*$ and $\bW^*$ have the top $Z$ eigenvectors of $\widetilde\bC_V(t)$ and $\widetilde\bC_V(0)$ respectively as their columns. To bound the difference between these two matrices, we can use the Davis–Kahan theorem. In particular, we invoke Theorem~2 of \cite{yu2015useful} which (when adapted to our setup) says that there exists an orthogonal matrix $\Tilde{\bO}$ such that:
\begin{align*}
    \norm{\widetilde\bW^*\widetilde\bO - \bW^*}_F \leq \frac{2^{\frac{3}{2}}\norm{\widetilde\bC_V(t) - \widetilde\bC_V(0)}_F}{\lambda_Z - \lambda_{Z+1}}
\end{align*}
Since $\bO$ can be any orthogonal matrix, let us choose $\bO = \Tilde{\bO}$. Then, we have:
\begin{align*}
     \norm{\bO^\top\bu - \Tilde{\bu}} &\leq \frac{c_1t}{\sqrt{M}} + \frac{2^{\frac{3}{2}}c_2 \log M}{\lambda_Z - \lambda_{Z+1}}\cdot \kappa \frac{t}{\sqrt{M}} \\
\end{align*}
where we have used (\ref{eqn:lem}). Therefore:
\begin{align*}
    \norm{\bO^\top\bu - \Tilde{\bu}} &\leq \zeta\frac{t}{\sqrt{M}}\left(1 + \frac{\xi \log M}{\lambda_{Z} - \lambda_{Z+1}} \right)
\end{align*}
where $\zeta = \frac{4\eta Q\sqrt{D} L_{\phi}^2c_{in}^2}{\delta}$ and $\xi = \frac{2^{\frac{7}{2}} QDc_{in}(L_{\phi}c_{\theta} + \abs{\phi(0)}) }{\delta}$ concluding the proof.
\end{proof}

\subsection{Proposition \ref{prop:eq_we}}
   Consider a neural network of the form (\ref{eqn: ntk_nn}) trained using a loss of the form (\ref{eqn:loss}) using gradient descent. Then for any input $\bx$,
    % \begin{align*} 
     $   {\bW_{.i}}(0) = {\bW_{.j}}(0)\Rightarrow f_i(\bx ; {\btheta(t)}) = f_j(\bx ; {\btheta(t)}),\  \forall \; t \geq 0.$
    % \end{align*}
\begin{proof}
We compare the outputs by analyzing the learning dynamics as given in Lemma~\ref{lemma:ntk_dyn}, we know that:
\begin{align*}
        \dot{f}_i({\bx}) &= \frac{-1}{N}\sum_{n, q} \frac{\partial l}{\partial s(\bx_n , \xpair )} \biggl[ \sum_{p=1}^{Z} \biggl[K_{ip}(\bx, {\bx_n}; \btheta)g_p(\bx_n , \xpair ) + K_{ip}(\bx, \xpair ; \btheta)g_p(\xpair , \bx_n ) \biggr]\biggr] \\
        \dot{f}_j({\bx}) &= \frac{-1}{N}\sum_{n, q} \frac{\partial l}{\partial s(\bx_n , \xpair )} \biggl[ \sum_{p=1}^{Z} \biggl[K_{jp}(\bx, {\bx_n};\btheta)g_p(\bx_n , \xpair ) + K_{jp}(\bx, \xpair ; \btheta)g_p(\xpair , \bx_n ) \biggr]\biggr]
    \end{align*}
where $g_p(\bx, \by) = \frac{\partial s(\bx, \by)}{\partial f_p(x;\btheta)}$. Note that irrespective of the similarity measure, the quantities $\frac{\partial l}{\partial s(\bx_n , \xpair )}$, $ g_p(\bx_n , \xpair )$, $g_p(\xpair , \bx_n )$ are independent of $i$ and $j$ and hence are identical in the dynamics of both $\dot{f}_i({\bx})$ and $\dot{f}_j({\bx})$.

\noindent Therefore, we only have to characterize and compare the NTK for a neural network $f(\cdot)$ of the form (\ref{eqn: ntk_nn}). Recall that the NTK can be defined as:
\begin{align*}
    \bK_{ij}(\bx, \by; \btheta) =  
    \underbrace{{\frac{\partial f_i(\bx; \btheta)}{\partial \bv}}^\top \frac{\partial f_j(\by; \btheta)}{\partial \bv} }_{\text{Term I}}
    + \underbrace{{\frac{\partial f_i(\bx; \btheta)}{\partial \bw}}^\top \frac{\partial f_j(\by; \btheta)}{\partial \bw}}_{\text{Term II}}
\end{align*}
where $\bv$ and $\bw$ are vectorized versions of weights $\bV$ and $\bW$ respectively. Now we can expand \emph{Term I}:
\begin{align*}
    {\frac{\partial f_i(\bx; \btheta)}{\partial {\bV_{qr}}}} &= \bW_{qi} \phi'({\bV_{q.}}\bx)\bx_r
\end{align*}
and therefore,
\begin{align*}
     {\frac{\partial f_i(\bx, \btheta)}{\partial \bv}}^\top \frac{\partial f_j(\by; \btheta)}{\partial \bv} &= \sum_{q, r} \bW_{qi} \bW_{qj} \phi'({\bV_{q.}} \bx) \phi'({\bV{q.}} \by) \bx_r \by_r.
\end{align*}
Similarly, for \emph{Term II}:
\begin{align*}
     {\frac{\partial f_i(\bx; \btheta)}{\partial {\bW_{q'r'}}}} &= \phi({\bV_{q'.}} \bx) \cdot \mathbf{1}_{i = r'},
\end{align*}
and,
\begin{align*}
    {\frac{\partial f_i(\bx; \btheta)}{\partial \bw}}^\top \frac{\partial f_j(\by; \btheta)}{\partial \bw} = \sum_{q'} \phi({\bV_{q'.}} \bx) \phi({\bV_{q'.}} \by) \cdot \mathbf{1}_{i = j}
\end{align*}
We are interested in the evolution of $f_i(\bx; \btheta(t))$ and $  f_j(\bx; {\btheta(t)})$ when ${\bW_{.i}}(0) = {\bW_{.j}}(0)$. Therefore,
\begin{align*}
    K_{ip}(\bx, {\bx_n}; \btheta) &= \sum_{q, r} \bW_{qi} \bW_{qp} \phi'({\bV_{q.}} \bx) \phi'({\bV_{q.}} \by) \bx_r \by_r +  \sum_{q'} \phi({\bV_{q'.}} \bx) \phi({\bV_{q'.}} \by) \cdot \mathbf{1}_{i = p} \\
    K_{jp}(\bx, {\bx_n}; \btheta) &= \sum_{q, r} \bW_{qj} \bW_{qp} \phi'({\bW_{q.}} \bx) \phi'({\bW_{q.}} \by) \bx_r \by_r +  \sum_{q'} \phi({\bV_{q'.}} \bx) \phi({\bV_{q'.}} \by) \cdot \mathbf{1}_{j = p}
\end{align*}
From these equations, it is clear that $\bK_{ip}(\bx, \bx_n; \btheta) =  \bK_{jp}(\bx, \bx_n; \btheta)$ whenever ${\bW_{.i}} = {\bW_{.j}}$. Therefore, if we prove that $\bW_{.i}(t) = {\bW_{.j}}(t)$ for all $t$, then $\dot{f}_i({\bx}) = \dot{f}_j({\bx})$ and hence ${f}_i({\bx}) = {f}_j({\bx})$ for all $t$.

\noindent Let us assume ${\bW_{.i}}(t) = {\bW_{.j}}(t)$ at some time $t$. Then, after an iteration of gradient descent, with learning rate $\eta$, we have:
\begin{align*}
    \bW_{.i} (t + 1) = \bW_{.i} (t) - \frac{\eta}{N}  \sum_{n=1}^{N} \sum_{q = 1}^{Q} {\frac{\partial l}{\partial s(\bx_k , \bx_{k,q})}} {\frac{\partial s(\bx_k , \bx_{k,q})}{\partial \bW_{.i} (t)}} \\
    \bW_{.j} (t + 1) = \bW_{.j} (t) - \frac{\eta}{N}  \sum_{n=1}^{N} \sum_{q = 1}^{Q} {\frac{\partial l}{\partial s(\bx_k , \bx_{k,q})}} {\frac{\partial s(\bx_k , \bx_{k,q})}{\partial \bW_{.j} (t)}}
\end{align*}
\subsubsection*{Case 1: Dot product}

We have:
\begin{align*}
    {\frac{\partial s(\bx_k , \bx_{k,q})}{\partial \bW(t)}} &= \frac{1}{M} {\frac{\partial  \text{Tr} 
 \left(\phi^\top (\bV(t)\bx_k )\bW(t)\bW(t)^\top \phi(\bV\bx_{k,q})\right)}{\partial \bW(t)}} \\
 &= \frac{1}{M} \left(\phi(\bV(t)\bx_{k,q})\phi^\top (\bV(t)\bx_k ) + \phi(\bV(t)\bx_k )\phi^\top (\bV(t)\bx_{k,q}) \right) \bW(t)
\end{align*}
Therefore:
\begin{align*}
    {\frac{\partial s(\bx_k , \bx_{k,q})}{\partial \bW_{.i} (t)}} &= \frac{1}{M} \left(\phi(\bV(t)\bx_{k,q})\phi^\top (\bV(t)\bx_k ) + \phi(\bV(t)\bx_k )\phi^\top (\bV(t)\bx_{k,q}) \right) \bW_{.i} (t) \\
    {\frac{\partial s(\bx_k , \bx_{k,q})}{\partial \bW_{.j} (t)}} &= \frac{1}{M} \left(\phi(\bV(t)\bx_{k,q})\phi^\top (\bV(t)\bx_k ) + \phi(\bV(t)\bx_k )\phi^\top (\bV(t)\bx_{k,q}) \right) \bW_{.j} (t)
\end{align*}
Clearly, since $\bW_{.i} (t) = \bW_{.j} (t)$, ${\frac{\partial s(\bx_k , \bx_{k,q})}{\partial \bW_{.i} (t)}} = {\frac{\partial s(\bx_k , \bx_{k,q})}{\partial \bW_{.j} (t)}}$ for all $n, q$ and therefore $\bW_{.i} (t + 1) = \bW_{.j} (t + 1)$.

\subsubsection*{Case 2: Cosine similarity}

From the proof of Lemma \ref{lemma:change}, we have:
\begin{align*}
    \frac{\partial s(\bx_k , \bx_{k,q})}{\partial \bW} 
    &= \frac{1}{\sqrt{M}}\left[\frac{\phi(\bV\bx_k )\bu_{k,q} ^\top  + \phi(\bV\bx_{k,q})\bu_k ^\top }{(\norm{\bu_k } + \delta) (\norm{\bu_{k,q} } + \delta)} \right]- \frac{1}{\sqrt{M}}\left[ \frac{\bu_k ^\top \bu_{k,q} }{(\norm{\bu_k } + \delta)^2( \norm{\bu_{k,q} } + \delta)^2} \right.\\
  &\quad\quad  \left.\left( \frac{(\norm{\bu_{k,q} } + \delta)\phi(\bV\bx_k ){\bu_k ^\top }}{ \norm{\bu_k }} + \frac{(\norm{\bu_k } + \delta)\phi(\bV\bx_{k,q}){\bu_{k,q} ^\top }}{ \norm{\bu_{k,q} }} \right) \right] 
\end{align*}
Therefore,
\begin{align*}
    \frac{\partial s(\bx_k , \bx_{k,q})}{\partial \bW_{.i} (t)} 
    &= \frac{1}{\sqrt{M}}\left[\frac{\phi(\bV\bx_k ) (u_{k,q})_i + \phi(\bV\bx_{k,q}) (u_{k})_i}{(\norm{\bu_k } + \delta) (\norm{\bu_{k,q} } + \delta)} \right] 
    - \frac{1}{\sqrt{M}}\left[ \frac{\bu_k ^\top \bu_{k,q} }{(\norm{\bu_k } + \delta)^2( \norm{\bu_{k,q} } + \delta)^2} \right.\\
    &\quad\quad\left.\left( \frac{(\norm{\bu_{k,q} } + \delta)\phi(\bV\bx_k ){ (u_{k})_i}}{ \norm{\bu_k }} + \frac{(\norm{\bu_k } + \delta)\phi(\bV\bx_{k,q}){ (u_{k,q})_i}}{ \norm{\bu_{k,q} }} \right) \right]
    \end{align*}
    and
    \begin{align*}
    \frac{\partial s(\bx_k , \bx_{k,q})}{\partial \bW_{.j} (t)} 
    &= \frac{1}{\sqrt{M}}\left[\frac{\phi(\bV\bx_k ) (u_{k,q})_j + \phi(\bV\bx_{k,q}) (u_{k})_j}{(\norm{\bu_k } + \delta) (\norm{\bu_{k,q} } + \delta)} \right] - \frac{1}{\sqrt{M}}\left[ \frac{\bu_k ^\top \bu_{k,q} }{(\norm{\bu_k } + \delta)^2( \norm{\bu_{k,q} } + \delta)^2}\right.\\
   & \quad\quad\left.\left( \frac{(\norm{\bu_{k,q} } + \delta)\phi(\bV\bx_k ){ (u_{k})_j}}{ \norm{\bu_k }} + \frac{(\norm{\bu_k } + \delta)\phi(\bV\bx_{k,q}){ (u_{k,q})_j}}{ \norm{\bu_{k,q} }} \right) \right] 
\end{align*}
Note that $ (u_{k,q})_i = \frac{1}{\sqrt{M}}{\bW_{.i}^\top }(t)\phi(\bV\bx_{k,q})$ and $ (u_{k,q})_j = \frac{1}{\sqrt{M}}{\bW_{.j}^\top }(t)\phi(\bV\bx_{k,q})$. Therefore, since $\bW_{.i} (t) = \bW_{.j} (t)$, $ (u_{k,q})_i =  (u_{k,q})_j $. Similarly, $ (u_{k})_i =  (u_{k})_j $. Since this is true for all $k, q$, $\bW_{.i} (t + 1) = \bW_{.j} (t + 1)$.

\noindent Therefore, for both dot product and cosine similarity based losses, $\bW_{.i} (t) = \bW_{.j} (t)$ implies $\bW_{.i} (t+1) = \bW_{.j} (t+1)$. Note that  $\bW_{.i} (0) = \bW_{.j} (0)$ is assumed to be true. Therefore, by the principle of induction,  $\bW_{.i} (t) = \bW_{.j} (t)$ for all $t \geq 0$. Hence, ${f}_i({\bx}) = {f}_j({\bx})$ for all $t \geq 0$ which concludes the proof.
\end{proof}

\clearpage
\section{Additional Plots}\label{app: additioanl plots}
We present in Figure~\ref{fig:addn_experiments} the same results as in Figure~\ref{fig:1layer},~\ref{fig:additional_experiments}(left)~\&~\ref{fig:PCA Theory}(left), but the change is plotted as a function of time instead of as a function of $M$. In the main paper, we analyze the behaviour of the quantities when $M$ changes, as this allows us to show that the change decreases with width. Here, we see that the increase with $t$ is also small for large widths. We also note that the standard deviation is significantly higher for smaller widths, which is to be expected since the bound is loose for small $M$. 
\begin{figure}[h]
\floatconts
  {fig:addn_experiments}% label for whole figure
  {\caption{ .
  (a) The maximum entry-wise change in empirical NTK for single hidden layer neural networks with ReLU activation and Linear loss  for varying width $M$. 
  (b) The maximum entry-wise change in empirical NTK for three hidden layer neural networks of varying width.
  (c) Change in $\bC_V$ during training for varying width $M$.
  }}% caption for whole figure
  {%
    \subfigure{%
      \label{fig: app 1layer_epoch}% label for this sub-figure
      \includegraphics[width=0.32\textwidth]{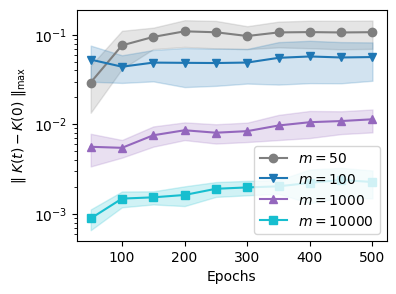}
    }
    \subfigure{%
      \label{fig: app 3layer_epoch}% label for this sub-figure
      \includegraphics[width=0.32\textwidth]{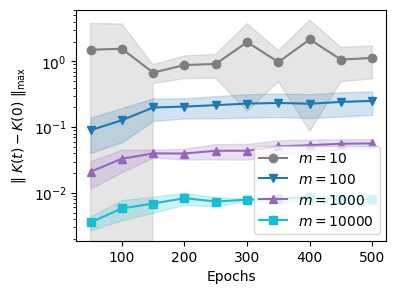}
    }%\qquad % space out the images a bit
    \subfigure{%
      \label{fig: app c_change_epoch}% label for this sub-figure
      \includegraphics[width=0.32\textwidth]{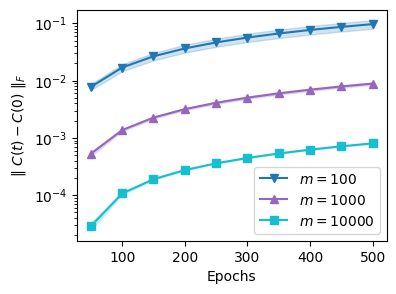}
    }%\qquad % space out the images a bit
  
  }
\end{figure}

\begin{figure}[h]
    \centering
    \includegraphics[width = 0.32\textwidth]{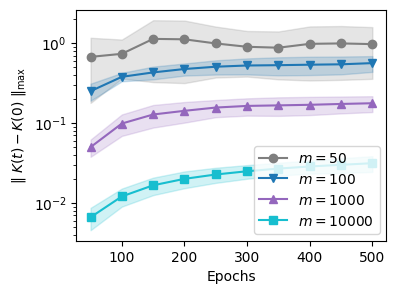}
    \caption{The maximum entry-wise change in empirical NTK for single hidden layer neural networks with \emph{Tanh} activation and Linear loss for varying depths $M$. }
    \label{fig: app tanh}
\end{figure}

\noindent In addition, we also observe that the evolution for  ReLU networks (Figure~\ref{fig:1layer}~\&~\ref{fig: app 1layer_epoch}) are  similar to the evolution observed for Tanh activation in Figure~\ref{fig: app tanh}.

\end{document}